\documentclass[a4paper,fleqn]{cas-dc}

\usepackage{hyperref}
\usepackage{url}

\usepackage{cite}
\usepackage[numbers]{natbib}
\usepackage{amsmath,amssymb,amsfonts}
\usepackage{stmaryrd}
\usepackage{algorithmic}
\usepackage{bm}
\usepackage{graphicx}
\usepackage{textcomp}
\usepackage{xcolor}
\usepackage{lipsum}
\usepackage{subcaption}
\usepackage{booktabs}
\usepackage{tablefootnote}
\usepackage{stfloats}
\usepackage{lastpage}
\makeatletter
\AtBeginDocument{\def\lastpage{\pageref*{LastPage}}}
\makeatother
\usepackage{tikz}
\usetikzlibrary{positioning, calc, chains, fit, shapes.geometric, arrows.meta, backgrounds}

\definecolor{compcolor}{RGB}{240, 249, 255}
\definecolor{opcolor}{RGB}{20, 100, 170}
\definecolor{gnncolor}{RGB}{40, 167, 69}
\definecolor{paramcolor}{RGB}{120, 120, 120}
\newcommand{\etalcite}[2]{#1~et~al.~\citep{#2}}

\def\tsc#1{\csdef{#1}{\textsc{\lowercase{#1}}\xspace}}
\tsc{WGM}
\tsc{QE}
\tsc{EP}
\tsc{PMS}
\tsc{BEC}
\tsc{DE}
\begin{document}
\let\WriteBookmarks\relax
\def\floatpagepagefraction{1}
\def\textpagefraction{.001}
\shorttitle{Evaluating Physics-Informed Graph Neural Networks for Robust Power Flow Prediction}
\shortauthors{E. Okoyomon et~al.}

\title [mode = title]{Evaluating Physics-Informed Graph Neural Networks for Robust Power Flow Prediction}

\author[1]{Ehimare Okoyomon}[orcid=0009-0004-1535-0967]
\cormark[1]
\fnmark[1]
% \ead{jkk@example.in}
% \ead[url]{www.jkkrishnan.in}

% \credit{Conceptualization of this study, Methodology, Software}

%\address[1]{, Street 129, 1043 NX Amsterdam, The Netherlands}
\affiliation[1]{organization={Technical University of Munich},
                country={Germany}}

\author[1]{Arbel Yaniv}
\author[1]{Christoph Goebel}

% \credit{Data curation, Writing - Original draft preparation}

\cortext[cor1]{Corresponding author}
\fntext[fn1]{Correspondence to: Ehimare Okoyomon \textless e.okoyomon@tum.de\textgreater.}

\begin{abstract}
Fast and reliable voltage prediction is critical for actively managed distribution grids. While Graph Neural Networks (GNNs) are reported to offer orders-of-magnitude speedups over classical solvers, their deployment is hindered by poor topological generalization. In this work, we systematically investigate three physics-informed inductive biases to assist in this learning task: physics-constrained loss functions, complex-valued neural networks, and residual-based task reformulation. To rigorously assess operational readiness, we depart from standard average-case metrics such as global RMSE, which can obscure extreme localized failures.
Using the heterogeneous ENGAGE dataset, we demonstrate that while our complex-valued GNN is the most accurate of the evaluated variants on known grids, with median errors within the tolerances of an illustrative smart-meter reference, severe out-of-distribution vulnerabilities remain across all architectures. For the evaluated GNN models and synthetic feeders, our findings show substantial cross-feeder errors, so that these models require application-specific validation and safeguards before operational deployment.
\end{abstract}

% % \includegraphics{figures/experiment_overview_diagram_physics-informed.pdf}

% \begin{highlights}
%     \item Design of complex-valued GNN (Cplx) for power flow prediction in distribution grids.
%     \item Cplx achieves the lowest median error of the GNN variants on familiar feeders.
%     \item Standard average error metrics inherently mask extreme localized failures.
%     \item Linearized baseline outperforms every GNN variant on familiar and unseen feeders.
%     \item Evaluated models require validation and safeguards before use on unseen feeders.
% \end{highlights}

\begin{keywords}
Distribution Grid Modeling \sep Generalization \sep Graph Neural Networks (GNNs) \sep Physics-Informed Machine Learning \sep Power Flow
\end{keywords}

\maketitle

\section{Introduction}

Distribution networks are transitioning from passive delivery systems into actively managed infrastructures. Driven by electrification and high penetrations of Distributed Energy Resources (DERs), modern grids operate under high volatility and tight real-time decision windows. In this environment, rapid and robust voltage prediction is a central task for maintaining equipment safety and system stability. Traditionally, solving the AC power-flow problem relies on classical iterative solvers to determine bus voltage magnitudes and phase angles from nodal injections. While accurate, these methods scale poorly and may fail to converge under realistic, heavily loaded conditions. Conversely, simplified linear approximations are computationally efficient but fundamentally unreliable for distribution networks, where high resistance-to-reactance (R/X) ratios cause large voltage drops and power losses, invalidating their core assumptions.

These computational bottlenecks have driven a surge of interest in machine learning approaches, particularly Graph Neural Networks (GNNs), which can achieve orders-of-magnitude speedups while retaining strong accuracy and flexibility \citep{donon_neural_2020,jeddi_physics-informed_2021}. Because GNNs naturally adapt to varying topologies, they are architecturally well-suited for the structural heterogeneity and frequent reconfigurations of distribution systems. However, purely data-driven models frequently fail to generalize to unseen grids. Compounding this generalization barrier is a critical flaw in how the research community evaluates model readiness, since standard global aggregate metrics, such as the mean Root Mean Square Error (RMSE), inherently mask severe, localized predictive failures. To transition ML solvers from academic novelties to operationally safe tools, we must systematically investigate how model design can bound these worst-case errors and drive models toward highly robust, topology-agnostic voltage predictions.

Inductive biases are assumptions embedded within a model's architecture, data representation, or training process to steer the learning algorithm toward structurally sound solutions. By capturing the underlying physics of the power flow problem, these biases are essential for bridging the gap between average-case benchmarks and the worst-case operational safety required for real-world deployment. While physics-informed approaches have gained traction by incorporating governing equations into the training regime, prior research has largely centred around equation-constrained losses. This narrow focus often overlooks alternative structural biases and relies on evaluation frameworks that fail to rigorously capture worst-case predictive fidelity. Ultimately, there remains a critical gap in understanding how distinct inductive biases empirically impact cross-system generalization which is an absolute necessity for deployment in distribution grids where unseen configurations are the norm.

\subsection{Contributions and Scope}

In this work, we rigorously investigate the role of physics-informed learning for voltage prediction in active distribution grids.
We analyze three classes of inductive biases:
\emph{Physics-constrained loss functions}, which incorporate the physical equations directly into the optimization objective;
\emph{Complex-valued neural networks}, which align the model's internal representations with the natural complex-valued structure of electrical quantities;
\emph{Task reframing as residual prediction}, where the model learns to approximate deviations from a known baseline rather than absolute voltages.
We claim novelty for none of these methodologies individually, but rather introduce comparability and more comprehensive evaluation for their use in power flow modeling.

For each method, controlled experiments are conducted across a dataset of common distribution grid topologies,
to assess the efficacy of current ML-based solvers for safety-critical deployment in modern distribution grids.
Due to their inherent adaptability to varying topologies, we focus strictly on GNN architectures.
In this light, we make the following core contributions\footnote{All models and code are publicly available at \url{https://github.com/EOkoyomon/pignn-pf-eval}}:

\begin{itemize}
    \item \textbf{Controlled Comparison of Biases:} We isolate three classes of physics-informed inductive bias under a fixed backbone, experiment setting, and training protocol. To our knowledge this is the first like-for-like comparison of these methods for power flow.
    \item \textbf{Complex-Valued GNN for Power Flow:} We introduce a complex-valued GNN for power flow prediction across multiple distribution feeders, distinguished from prior complex-valued work by task, architecture, and evaluation setting.
    \item \textbf{Exposure of Operational Limitations:} We systematically assess cross-system generalization and tail behaviour of these neural solvers and reveal failures that standard evaluations miss; these failures currently place the models outside required reliability margins.
\end{itemize}

\section{Background}

\subsection{Role of Inductive Biases in Machine Learning}

Inductive biases are the foundational assumptions embedded in a machine learning model or algorithm. They encode prior knowledge, constraints, or assumptions to guide how a model generalizes from training data to unseen samples. Thus, the design and selection of inductive biases allow models to perform well in varied settings, boosting generalization and interpretability, while also imposing trade-offs between efficiency and flexibility. These biases are essential when learning from finite data, as they help constrain the hypothesis space and enable effective learning. Without assumptions, there are infinitely many functions that can fit the training data equally well, so machine learning models need a way to prioritize certain hypotheses over others.

However, inappropriate bias can lead to poor performance if the assumptions do not match the underlying data distribution. This highlights the importance of carefully selecting and designing inductive biases to suit the specific problem and data at hand.

\subsection{Physics-Informed ML in Power Systems}

Physics-informed (PI) machine learning integrates domain knowledge into machine learning models to enhance their performance, interpretability, and reliability \citep{karniadakis2021physics}. This approach is particularly relevant in engineering domains like power systems, where physical principles govern system behaviour. By embedding these principles into the learning process, this technique aims to provide more trustworthy models that can generalize better to unseen scenarios. In power systems, physics-informed neural networks (PINN) have been applied to various tasks, including state estimation, dynamics analysis, and optimal power flow \citep{huang_applications_2023}. Although these approaches claim to improve model generalization, there have been no dedicated generalization studies of these methods for power systems, thus motivating our systematic assessment. In the following, we review three strategies of physics-informed machine learning that are particularly relevant to our study of voltage prediction in distribution grids. 

We do not claim this taxonomy exhausts physics-informed modeling, but rather use these to isolate three points at which a physics prior can enter a learning system: the \emph{objective}, which determines what is penalized; the \emph{representation}, which determines what the hypothesis space can express; and the \emph{task parameterization}, which determines what the model is asked to predict. The three biases studied here are canonical power systems instances of each entry point, namely the AC power-balance penalty, the complex-valued representation, and the slack-referenced residual target, respectively. Because the three act at different points in the pipeline, each can be introduced or removed independently while holding the backbone, data, and optimizer fixed.

\subsubsection{Physics-Constrained Loss Function}

Constraining loss functions is one of the most common approaches to physics-informed machine learning in power systems and was highlighted as a core PINN paradigm by a recent review paper \citep{huang_applications_2023}. This is typically achieved by assigning physical meanings to the output variables of the neural network and embedding the governing equations of these physical variables as an additional loss term. The loss function can thus be formally written as:

\begin{equation}
    \mathcal{L} = L_{\text{pred}}(\hat{\mathbf{y}}, \mathbf{y}) + \gamma L_{\text{reg}}(\mathbf{W}) + \lambda L_{\text{phys}}(\mathbf{X}, \hat{\mathbf{y}})
\end{equation}

where $L_{\text{pred}}$ is a typical loss function, such as Mean Squared Error (MSE), which measures the difference between the predicted and target output ($\hat{y}$ and $y$ respectively); $L_{\text{reg}}$ is a regularization term (such as L1 or L2 norm) imposed on the weights $W$ of the neural network; and $L_{\text{phys}}$ is a differentiable, physical regularization term that quantifies the degree to which the model's predictions violate the governing physical equations, with $X$ representing the input features.
Note that $L_{\text{phys}}$ does not depend on $y$, allowing the network to be trained in an unsupervised manner. $\gamma$ and $\lambda$ are thus hyperparameters that balance the contributions of the different loss components.

For power flow estimation, $L_{\text{phys}}$ is generally composed of the power balance equations of the AC power flow formulation.
More specifically, using $P$ as the input active power, $Q$ as the input reactive power, $\hat{V}$ as the predicted voltage magnitude, and $\hat{\theta}$ as the predicted voltage angle, we can express the AC power flow loss term as in Equations~\ref{eq:active_power}--\ref{eq:reactive_power},
where $i$ is a particular bus in an $N$-bus network, and $G_{ik}$ and $B_{ik}$ are the real and imaginary parts of the $ik$-th element of the network's nodal admittance matrix, respectively.
These nodal losses can then be combined to provide a physics-constrained loss for a particular network sample, commonly formulated as a composite mean squared error for a graph $G$ with $N_G$ nodes, such as in Equation~\ref{eq:physics_loss}.

\begin{samepage}
    \small
    \begin{equation}
        L_{P_i} = P_i - \sum_{k=1}^N |\hat{V_i}| |\hat{V_k}| \left( G_{ik} \cos(\hat{\theta_i}-\hat{\theta_k}) + B_{ik} \sin(\hat{\theta_i} - \hat{\theta_k}) \right) \label{eq:active_power} \\
    \end{equation}
    \begin{equation}
        L_{Q_i} = Q_i - \sum_{k=1}^N |\hat{V_i}| |\hat{V_k}| \left( G_{ik} \sin(\hat{\theta_i}-\hat{\theta_k}) - B_{ik} \cos(\hat{\theta_i} - \hat{\theta_k}) \right) \label{eq:reactive_power}
    \end{equation}
\end{samepage}

\begin{equation}
    % \small
    L_{\text{phys},G} = \frac{1}{N_G}\sum_{i=1}^{N_G} (L_{P_i}^2 + L_{Q_i}^2) \label{eq:physics_loss}
\end{equation}

Since this method applies to the training process itself, it is model-agnostic and can be adapted for arbitrary neural network architectures. 

\subsubsection{Complex-Valued Neural Networks}

Complex-valued neural networks (CVNNs) extend traditional real-valued neural networks by allowing weights, biases, and activations to be complex numbers. This enables the model to capture phase relationships and interactions between real and imaginary components more naturally, potentially leading to improved performance in tasks involving complex-valued data. CVNNs have been shown to be effective in various domains, including computer vision, speech and signal processing, medical image reconstruction, and control systems \citep{lee2022complex}.

A typical layer computes
\[
z' = W z + b, \quad W \in \mathbb{C}^{m \times n}, \; z \in \mathbb{C}^n,
\]
which in real--imaginary form expands as
\[
z' = (W_r x - W_i y) + i (W_r y + W_i x), \quad (z = x + i y).
\]
Nonlinearities are either applied directly component-wise, \(\sigma(z) = \sigma(x) + i \sigma(y)\), or in polar form, \(\sigma(z) = f(|z|) e^{i g(\angle z)}\), where $f$ and $g$ are nonlinear functions applied to the magnitude and phase, respectively. Training employs \emph{Wirtinger derivatives}, which calculate partial derivatives based on the complex weights and their complex conjugates, allowing standard gradient-based optimization in \(\mathbb{C}\). Normalization and loss functions are defined analogously to the real case, often by acting on real and imaginary parts separately or jointly via \(|z|\). 
In power systems, CVNNs naturally accommodate phasor quantities such as complex voltages and power flows. This allows voltage magnitudes and angles to be modeled and predicted jointly in power flow, preserving their inherent algebraic structure rather than separating them as real-valued models do.

\subsubsection{Residual Prediction}

Residual prediction reframes the learning task from predicting absolute values to predicting deviations from a known baseline or approximate solution, which can simplify the learning problem. The approach has been popularized in models such as ResNet \citep{he2016deep} and XGBoost \citep{chen2016xgboost}, which have had significant impacts in domains like computer vision and time-series forecasting. In physics-informed machine learning, 
% scientific computing, 
exploiting a known approximate solution is a common technique, and has been shown to reduce both error and predictive uncertainty relative to purely data-driven models, for example when predicting airfoil lift \citep{pawar2021physics}.
Formally, whereas a standard regression model learns a map from the inputs directly to the absolute targets, $\hat{\mathbf{y}} = f_{\theta}(\mathbf{X})$, residual prediction presumes a cheaply computable baseline $\mathbf{y}_0(\mathbf{X})$ and restricts the learned map to the correction of that baseline, as in Equation~\ref{eq:residual_pred}, so that the network is fitted against the residual targets $\Delta\mathbf{y} = \mathbf{y} - \mathbf{y}_0(\mathbf{X})$ rather than against $\mathbf{y}$.
\begin{equation}
    \hat{\mathbf{y}} = \mathbf{y}_0(\mathbf{X}) + f_{\theta}(\mathbf{X}), \qquad
    L_{\text{pred}} = \left\lVert \Delta\mathbf{y} - f_{\theta}(\mathbf{X}) \right\rVert^2
    \label{eq:residual_pred}
\end{equation}

Since $\mathbf{y}_0$ is fixed and independent of the model parameters $\theta$, the baseline contributes no gradient during training, and the entire capacity of the model is therefore directed at the component of the signal that the baseline fails to explain.
This does not make the deviation itself easier to learn, but rather re-centres the targets near zero and directs model capacity at the component that is difficult.
Due to the complex physics of power systems, machine learning training can be unstable and heavily reliant on initializations and hyperparameters. However, voltages are generally close to a constant baseline set by the slack bus, thus motivating the use of residuals:
by focusing on capturing deviations, the redundant learning of the trivial baseline is eliminated.

\begin{table*}[t]
\centering
\caption{Selected physics-informed neural network studies for power flow. Adjacent tasks are included where the bias is transferable.
\emph{Tail metrics} reports whether worst-case or tail behaviour is presented alongside average-case error. Tasks: PF~--~power flow, SE~--~state estimation, OPF~--~optimal power flow, DyR~--~dynamic reconfiguration, SP~--~storage planning, PE~--~parameter estimation.}
\label{tab:pinn_overview}
\footnotesize
\setlength{\tabcolsep}{3pt}
\renewcommand{\arraystretch}{1.15}
\begin{tabular}{@{}l >{\raggedright\arraybackslash}p{1.4cm} >{\raggedright\arraybackslash}p{1.75cm} >{\raggedright\arraybackslash}p{0.95cm} >{\raggedright\arraybackslash}p{1.90cm} >{\raggedright\arraybackslash}p{1.95cm} >{\raggedright\arraybackslash}p{2.35cm} >{\raggedright\arraybackslash}p{2.45cm}@{}}
\toprule
\textbf{Study} & \textbf{Bias} & \textbf{Model} & \textbf{Task} &
\textbf{Grid level} & \textbf{Test system} &
\textbf{Topology Generalization} & \textbf{Tail metrics} \\
\midrule

\citet{donon_neural_2020}
  & Loss & Custom GNN & PF & Transmission
  & IEEE 9, 14, 30, 118
  & Cross-system
  & None \\
  \addlinespace[3pt]

\citet{hu2020physics}
  & Loss & MLP autoencoder & PF & Transmission
  & IEEE 57, 118
  & None
  & None \\
  \addlinespace[3pt]

% \citet{jeddi_physics-informed_2021}
\etalcite{Jeddi}{jeddi_physics-informed_2021}
  & Loss & GAT & PF & Transmission
  & IEEE 9, 14, 30, 118
  & Cross-system
  & None \\
  \addlinespace[3pt]

\citet{bottcher_solving_2023}
  & Loss & Recurrent GNN & PF & Distribution
  & Synthetic MV, SimBench MV, real MV
  & Synthetic $\rightarrow$ Simbench/real
  & None \\
  \addlinespace[3pt]

\citet{Habib_Isufi_vanBreda_Jongepier_Cremer_2024}
  & Loss & GNN & SE & Distribution
  & IEEE 14, 70, 179
  & None
  & None \\
  \addlinespace[3pt]

\citet{authier2024physics}
  & Loss & GNN & DyR & Distribution
  & BW-33, TPC-94
  & Switch configurations
  & Max.\ constraint violation \\
  \addlinespace[3pt]

% \citet{nellikkath2022physics}
\etalcite{Nellikkath}{nellikkath2022physics}
  & Loss & MLP & OPF & Transmission
  & IEEE 14, 39, 118, 162
  & None
  & Max constraint violation \\
  \addlinespace[3pt]

\citet{motalleb2016optimal}
  & Complex & MLP & SP & Transmission
  & 184 bus island
  & None
  & None \\
  \addlinespace[3pt]

\citet{wang2024joint}
  & Complex & CV-GCN & PE & Distribution
  & IEEE 33, 69
  & None
  & None \\
  \addlinespace[3pt]

\citet{ceylan2005comparison}
  & Complex & MLP ($\leq 8$ neurons) & PF & Toy
  & 6-bus network
  & None
  & None \\
  \addlinespace[3pt]

\citet{wu2023complex}
  & Complex & Cplx-STGCN & SE & Transmission
  & IEEE 118
  & N-1
  & None \\
  \addlinespace[3pt]

% \citet{chen2023physics}
\etalcite{Chen}{chen2023physics}
  & Residual & MLP & PF & Transmission
  & IEEE 30, 118, 300, SouthCarolina-500, PEGASE-1354
  & None
  & None \\
  \addlinespace[3pt]

\citet{za2027residual}
  & Residual & GNN w/ attention & OPF & Transmission
  & IEEE 57, 18, 2000
  & N-1
  & Max constraint violation \\
  \addlinespace[3pt]

\textbf{This work}
  & \textbf{Loss, complex, residual}
  & \textbf{GraphConv MPNN} & \textbf{PF} & \textbf{Distribution}
  & \textbf{ENGAGE (LV/MV feeders)}
  & \textbf{Leave-one-feeder-out}
  & \textbf{Max error + IQR, per-scenario} \\

\bottomrule
\end{tabular}
\end{table*}

\section{Related Works}

An overview of relevant works is presented in Table \ref{tab:pinn_overview}.

\paragraph{Physics-Constrained Loss Function.} Due to the ubiquity of this approach, we only highlight the most relevant studies here. GraphNeuralSolver \citep{donon_neural_2020} was one of the first works to apply this technique for power flow estimation, using their own custom GNN architecture and serving as a benchmark for many future works in this direction. \cite{hu2020physics} similarly applies this loss regularization technique to an MLP model to show that they are able to more robustly handle unfamiliar load profiles.
\citep{jeddi_physics-informed_2021} integrates this method into a Graph Attention Network (GAT) model in order to learn the importance of neighbouring nodes, while \citep{bottcher_solving_2023} adopts a randomized, recurrent GNN architecture, so that its predictor learns how to solve the power flow problem from many different starting points. All three works solve the problem in a strictly unsupervised manner, completely skipping the need for labeled training data and the $L_{\text{pred}}$ term. Other notable works include \citep{Habib_Isufi_vanBreda_Jongepier_Cremer_2024}, \cite{authier2024physics}, and \cite{nellikkath2022physics} who successfully apply this technique to distribution grid state estimation, dynamic reconfiguration, and optimal power flow.

\paragraph{Complex-Valued Neural Networks.} To date there are relatively few works that explore complex-valued neural networks for power system applications, despite the fact that electrical quantities such as voltages and currents are inherently complex-valued. 
A complex-valued network was used by \citet{motalleb2016optimal} as a component within a storage siting procedure to forecast bus voltage magnitudes and angles in a temporally autoregressive setting on a 184-bus island system, where past voltages drive the prediction. More closely related to the distribution setting, \citet{wang2024joint} propose a complex-valued graph convolutional network that embeds the AC power flow equations directly in the convolution operator. Their task, however, differs from ours: voltages are taken as inputs and the network solves for the topology and line parameters, whereas we estimate the complex nodal voltages given the injections and a known network.
\citep{ceylan2005comparison} makes an early attempt at a CVNN for load flow prediction and noted improved reliability compared to a real valued neural network baseline. However, both models were very small MLPs, consisting of one hidden layer of maximum 8 neurons, and evaluation was performed on a toy 6-bus network. Thus, the results seem hardly tractable for a real-world setting. A more comprehensive study is presented in \citep{wu2023complex}, which introduces a complex-valued spatio-temporal graph convolutional network (Cplx-STGCN) for the task of power system state estimation (PSSE) and highlights the advantages of this model against several baselines. However, their evaluation is limited to a single IEEE 118-bus transmission system and their performance increase cannot be attributed to the complex-valued network alone, as their CplxFNN model performs poorly and they do not evaluate real-valued counterpart of Cplx-STGCN. Furthermore, the PSSE task is different from our setting as it aims to predict unknown voltages in a network using historical voltages and a set of available measurements (effectively an interpolation problem), while our voltage prediction task aims to estimate complex nodal voltages given only complex power measurements,
effectively learning the power balance equations. To our knowledge, this study contributes the first evaluation of complex-valued neural networks for power flow across multiple realistic distribution feeders, to assess whether their reported benefits persist under out-of-distribution conditions.

\paragraph{Residual Prediction.} Currently, state-of-the-art power flow methods typically predict the absolute voltage targets directly.
Residual learning was employed in \citep{chen2023physics} for probabilistic power flow
and simulated on standard transmission system test cases. 
Beyond voltage prediction, \citep{za2027residual} uses residual learning  to learn the nonlinear residual between a given DC-OPF solution and the corresponding AC-OPF solution. 
When compared against models that directly learn the AC-OPF solution, the residual-learning framework achieved a 35–45\% reduction in voltage, power, and feasibility errors, demonstrating its effectiveness for OPF prediction. The success of this approach suggests that residual learning may also be advantageous for voltage prediction.
\section{Methods}

In this study we train models to perform \textbf{AC power flow prediction}.
The objective of AC power flow analysis is to determine the active power, reactive power, voltage magnitude, and voltage angle of every bus in a power network. For every bus, there is a set of known and unknown parameters, differing based on bus type, 
and the task is to estimate the unknowns. In distribution grid voltage prediction, our main task is to predict the voltage magnitude and angle of every PQ bus in the network, given PQ bus power injections and slack bus voltages. The solutions to the equations are the values that satisfy equations \ref{eq:active_power}–\ref{eq:reactive_power} when $L_{P_i} = L_{Q_i} = 0$.

\subsection{Model Design}\label{sec:model_design}

We provide a \texttt{Base} GNN architecture to serve as a control model for our experiments. We employ a GraphConv-based message-passing architecture with batch normalization, following the standard GNN framework established by \citet{morris2019weisfeiler}. This provides strong representational capacity while maintaining theoretical grounding in GNN expressivity research.
As input, the model accepts the active and reactive power injections and topological distance to slack bus as node features, and the line resistances and reactances as edge features. Additionally, every graph has access to the slack bus information (slack voltage magnitude, angle, resistance, and reactance) as global features. The model's task is then to predict the voltage magnitude and angle at every bus.
The model utilizes two-layer preprocessing and postprocessing stages, implemented using fully connected networks with ReLU activations and batch normalization. The core processing consists of seven 128-dimensional message-passing layers using the GraphConv update scheme. This model configuration was chosen via experimentation and hyperparameter tuning of previous paper architectures \cite{okoyomon2025framework, hansen_power_2022} and we found that our model enhancements improved performance in this distribution grid setting.

For all models, we keep the architecture as similar as possible to the baseline, only changing the components necessary to implement each inductive bias. This allows us to isolate the effects of each bias while minimizing confounding factors. 

\subsection{Power Flow Loss Model (\texttt{PFLoss})}

We implement a loss function that combines the standard supervised loss with a physics-based regularization that penalizes violations of the AC power flow equations. More specifically, we train the baseline GNN as usual, and then fine-tune the model using a weighted physics-constrained loss. This scheme was deliberately chosen to ensure this phase acts solely as a structural refinement, preventing it from becoming a deep re-optimization of the landscape. 
Using Equation \ref{eq:physics_loss} as a reference, we begin with $\lambda=0$ for the first phase, making this training completely supervised, before then setting $\lambda = 10^{-8}$ in the fine-tuning stage to incorporate the unsupervised physics loss. 
This scheme proved to be much more reliable than training completely unsupervised or with only a weighted physics loss, since these methods are highly reliant on hyperparameters and drastically increase training time with marginal benefits.
For example, training using a completely unsupervised regime resulted in longer, more unstable convergence, as well as errors at least two orders of magnitude worse than our fine-tuned model (see Appendix~\ref{app:pf_loss_training_ablation}).
This experience is aligned with the findings of other studies that reveal the fundamental challenge of training using a physics-constrained loss, particularly around the inherent sensitivity when balancing the conflicting objectives of the data (i.e. supervised) and physics (i.e. unsupervised) terms \citep{becerra2026role}.

To train the PFLoss model, we use the mean nodal power mismatch as our loss, similar to the methods of \citep{jeddi_physics-informed_2021} and \citep{donon_neural_2020}. This technique resulted in smoother gradients and better convergence when compared to using the sum of nodal mismatches or log of the sum such as in \cite{bottcher_solving_2023}. When training on heterogeneous networks, using a mean nodal loss 
also allows us to avoid
unfairly prioritizing larger networks with more nodes during training. 
Using Equation \ref{eq:physics_loss}, our physical regularization term becomes: 
\begin{equation}
    % \small
    L_{\text{phys}} = \frac{1}{|\mathcal{G}_j|} \sum_{G \in \mathcal{G}_j} L_{\text{phys},G}
\end{equation}
Per graph, $L_{\text{phys},G}$ is implemented as the mean of the $2N_G$ squared active and reactive residuals of the PQ buses, and the fine-tuning objective is $\mathcal{L}_{\text{NMSE}} + \lambda L_{\text{phys}}$. The residuals are computed from the admittance matrix and power injections of the pandapower model in per unit on its $1$~MVA system base, numerically equal to MW and Mvar, without feeder-specific normalization.
Importantly, since we only predict voltages for PQ buses, and treat slack injections as unknown, we only backpropagate on non-slack node mismatches. Since the data is derived using pandapower \citep{thurner2018pandapower}, we replicate their AC power balance equations and model assumptions to implement our loss function. We ensure that the power flow loss function provides negligible power mismatch values for the ground truth voltage targets before using it for model training.

\subsection{Complex-Valued Model (\texttt{Cplx})}\label{sec:cplx}

The \texttt{Cplx} model retains the architecture of Section~\ref{sec:model_design} with each real-valued operation replaced by its complex counterpart, so that it differs from the \texttt{Base} model in representing its internal features as phasors. To accommodate this representation, it also uses a phasor output and a phasor loss, and we evaluate these together as a single design rather than attributing its effect to any one component.
Each message-passing layer applies the aggregation with complex linear maps in place of real ones, i.e.:

\begin{equation}
    \mathbf{h}_i' = W_{\text{root}}\,\mathbf{h}_i + W \sum_{j \in \mathcal{N}(i)} e_{i,j}\,\mathbf{h}_j ,
    \qquad W, W_{\text{root}} \in \mathbb{C}^{m \times n} \nonumber
\end{equation}

where a complex linear map expands as $Wz = (W_r x - W_i y) + j(W_r y + W_i x)$ for $z = x + jy$. Nonlinearities use $\mathbb{C}$ReLU, applying ReLU to real and imaginary parts independently, and normalization applies independent batch normalization to each part.

The model predicts the complex voltage phasor $\hat{V}_i \in \mathbb{C}$ at each bus, a single complex value per node. Consistent with this phasor output, we minimize the mean squared phasor error $\frac{1}{N}\sum_i |\hat{V}_i - V_i|^2$, coupling magnitude and angle through the phasor geometry.
Targets are formed from the ground-truth polar quantities as $V_i = |V_i|e^{j\theta_i}$ with $\theta_i$ in radians, so magnitude and angle are the modulus and argument of one physical phasor rather than two independent real channels. 
Predictions are converted for evaluation as $|\hat{V}_i|$ and $\theta_i = \arg(\hat{V}_i)$, the latter reported in degrees. Because the angle is recovered as the argument of a phasor rather than regressed directly, it is bounded to $(-\pi, \pi]$ by construction; angular errors for all models are evaluated with the circular difference defined in Section~\ref{sec:metrics}. Full implementation details are given in Appendix~\ref{app:cplx}.

\subsection{Residuals Model (\texttt{Res})}

We implement a GNN architecture that predicts the residuals as deviations from the slack bus voltage. The model thus only needs to learn the differences from a known baseline, which are often smaller in magnitude and may exhibit simpler patterns than the original target variable. We implement the residuals model by enforcing a single skip connection that connects the neural network's output to the slack bus reference value \citep{donti2017task}. This way, the model learns to output residuals, which are added to the slack bus reference values to produce the final voltage predictions. 
This allows us to keep the input data format consistent across all models, simplifying data handling and preprocessing.

\subsection{Additional Baselines}\label{sec:methods_baselines}

The primary goal of deploying neural solvers is to provide high-speed predictions while remaining significantly more accurate than fast, simplified linear approximations. We thus introduce two commonly used linear approximations of the AC power flow solution as baseline comparators: LinDistFlow (LDF) and DC Power Flow (DCPF). LinDistFlow is a standard linear approximation specifically tailored to the high $R/X$ ratios and reactive power dynamics typical of radial distribution systems. However, its core assumptions rely on radial topologies and predominantly unidirectional power flow. To account for the high DER penetration and the meshed configurations present in the Medium-Voltage (MV) datasets, we also include DC Power Flow as a secondary baseline. While DCPF is typically reserved for transmission networks due to its assumption of negligible line resistance, we include it here as a topology-agnostic analytical comparison that can handle meshed networks. The primary advantage of these models in this context is that they do not need to be trained; thus, performance is independent of training data splits. We present these baselines not for their sophistication, but rather as established benchmarks for evaluating generalization failure, consistent with prior studies \citep{hansen_power_2022, yaniv_towards_2023, okoyomon2025framework}. These works observe that while GNN-based power flow methods perform exceptionally well on known configurations, they frequently falter on unseen topologies, ultimately being outperformed by rudimentary approximations like DCPF. Detailed formulations for both baselines can be found in the supplemental materials. LDF is applied along the shortest-hop spanning tree from the slack bus, disregarding any branches that close a loop, and models transformers with their tap ratios and phase shifts. These transformer parameters are not part of the GNN inputs described in Section~\ref{sec:model_design}, so that differences in transformer configuration between feeders are not observed by the learned models, which is most relevant on held-out feeders.

To further improve interpretability, we compare results against a simulated smart meter that perturbs the true state within the stated tolerances of a widely deployed metering device, $1\%$ ($0.01$ p.u.) for voltage magnitude and $1^{\circ}$ for phase angle \citep{IMSys_ref}. We use a meter rather than a phasor measurement unit (PMU) as the reference because PMUs remain too sparsely deployed to serve as a measurement basis at distribution voltage levels, where advanced metering infrastructure is the prevalent source \citep{cheng2023survey}. We adopt these values as an illustrative reference for the accuracy of instrumentation currently available in distribution networks rather than as a standards-derived threshold. In particular, the meter's angle tolerance concerns a locally measured phase relationship rather than a bus voltage angle referenced to the slack, so that the $1^{\circ}$ reference applies to the predicted angles only illustratively. Our simulated meter is an idealized reference, perturbing the true state with independent noise alone, whereas deployed metering infrastructure is additionally subject to latency, missing, and corrupted values. See Appendix~\ref{app:sensor} for details.

\section{Experiments}

\subsection{Experiment Dataset}

\begin{figure*}
    \centering
    \includegraphics[width=0.85\linewidth]{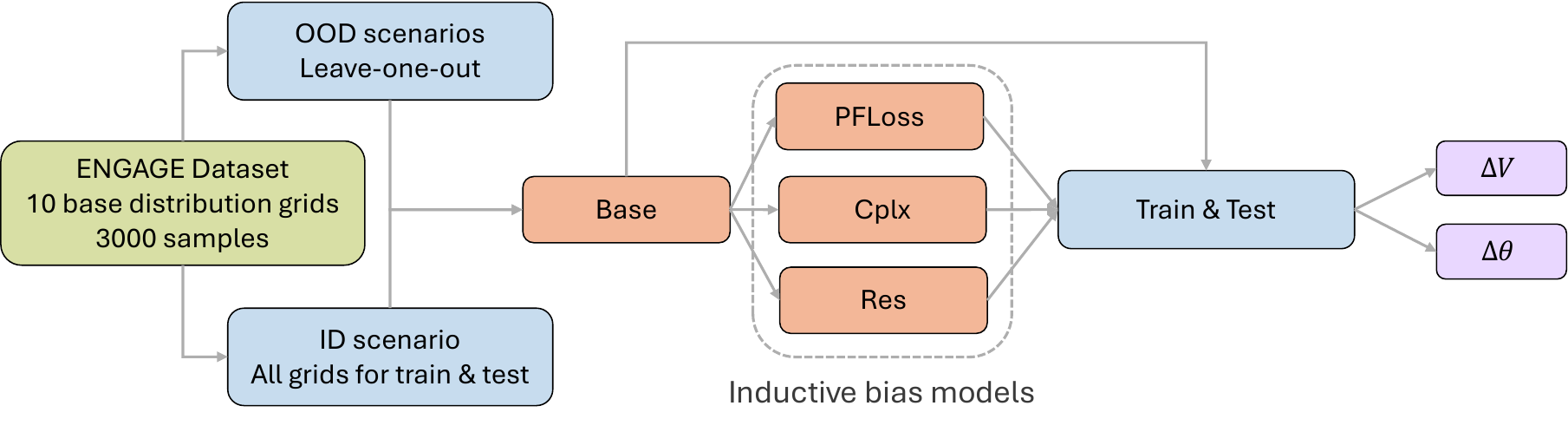}
    \caption{Experiment methodology for in-distribution and out-of-distribution model evaluation}
    \label{fig:experiment_overview}
\end{figure*}

We evaluate all models across a range of distribution networks. For controlled comparison of the three inductive biases, we conduct a series of experiments using a consistent dataset, model architecture, and evaluation metrics. The experimental overview is presented in Figure \ref{fig:experiment_overview}. We utilize the ENGAGE dataset \citep{okoyomon_2025_15464235}, which is a collection of low and medium voltage grids based on the SimBench networks \citep{meinecke_simbenchbenchmark_2020}. This test suite was introduced in \citep{okoyomon2025framework} to evaluate generalization capabilities of power flow models across several distribution grid configurations and promote robust grid planning and operation. With 3000 sample networks derived from 10 base distribution grids, the dataset contains test cases with varying sizes, topologies, and electrical characteristics, making it ideal for evaluating generalization performance. 
An overview of the experiment dataset including per-grid statistics and generation procedure is found in Appendix \ref{app:experiment-setting}.

\subsection{Evaluation Procedure} \label{sec:evaluation_procedure}

We first establish a baseline to identify how well each model performs under known network configurations and learned data distributions. This is the \textit{in-distribution (ID)} experiment group. We randomly shuffle the data with all 3000 grids and adopt a 75/15/10 train/validation/test split for model training. To evaluate generalization performance, we conduct a suite of \textit{out-of-distribution (OOD)} experiments. We use a leave-one-out approach where we train on all but one of the base grid configurations and test on the held-out configuration. The nine training feeders are split randomly into $2250$ training and $450$ validation samples, and all $300$ samples of the held-out feeder form the test set, compared to $2250$/$450$/$300$ samples drawn from all ten feeders in the ID setting. This approach allows us to assess how well each model can generalize to unseen grid topologies and operating conditions, which is critical for real-world deployment. Though a one-to-one grid evaluation is possible, previous studies indicate that current ML solvers are not mature enough for this use case \citep{bottcher_solving_2023, hansen_power_2022, jeddi_physics-informed_2021, okoyomon2025framework}. As such, we focus on the OOD setting for our generalization study. Therefore, we have 11 test scenarios in total: one for the ID grids scenario and 10 for the OOD scenarios. 

For all models, we apply uniform training procedures to ensure a fair comparison, using 4000 epochs and batch sizes of 128. The real-valued variants use the Normalized MSE ($\mathcal{L}_{\text{NMSE}}$) as the supervised loss function, while \texttt{Cplx} uses the phasor loss of Section~\ref{sec:cplx}. $\mathcal{L}_{\text{NMSE}}$ is a variant of the standard MSE loss that rescales the squared error in each feature dimension $d$ by the inverse of that dimension's L2 norm (computed across the batch):
\begin{equation}
    \mathcal{L}_{\text{NMSE}} = \frac{1}{ND}\sum_{d=1}^{D}\sum_{n=1}^{N} \frac{\left(\hat{y}_{n,d} - y_{n,d}\right)^2}{\|y_d\|_2 + \epsilon}
    \label{eq:nmse}
\end{equation}

where $\|y_d\|_2$ is the unsquared L2 norm of the targets in dimension $d$ over the $N$ PQ nodes of the batch, with voltage magnitude in p.u., voltage angle in degrees, and $\epsilon = 10^{-8}$.
This gives features with small magnitudes comparable influence on the loss to features with large magnitudes, preventing large-scale dimensions from dominating the gradient. We use an Adam optimizer with an adaptive learning rate starting from $10^{-4}$ until $10^{-5}$, reducing by half after 200 epochs, with a total patience of 750 epochs on the validation set. In each scenario, the learning-rate schedule and early stopping use only that scenario's validation split. Using a grid-search hyperparameter tuning over layers, learning rate, hidden dimensions, batch size and epoch budget (Appendix~\ref{app:hyperparameters}), we determine the optimal model configuration and training pipeline to ensure strong performance on standard accuracy metrics before proceeding with the experiment variants. 
We run all experiments over five random seeds ([12, 47, 34, 59, 73]) to account for training variability.

\subsection{Performance Metrics}\label{sec:metrics}

To rigorously assess the models, we depart from relying solely on standard global averages like Root Mean Squared Error (RMSE). While RMSE is common in machine learning literature, it implicitly blends typical predictive errors with heavy outlier penalties, which can obscure true operational performance.
Instead, we establish an evaluation framework designed to validate the accuracy range of the state predictions in a variety of distribution grid settings, emphasizing out-of-distribution (OOD) generalization.
Formally, we assess the absolute nodal error ($E_i$) for voltage magnitude ($E_i = |V_i - \hat{V}_i|$) and voltage angle ($E_i = |\operatorname{wrap}(\theta_i - \hat{\theta}_i)|$), where $\operatorname{wrap}(\cdot)$ maps angular differences into $[-180^{\circ}, 180^{\circ})$ so that, for example, $179^{\circ}$ and $-179^{\circ}$ are separated by $2^{\circ}$. We compare these errors against the tolerances of a widely deployed smart meter, $1\%$ for voltage magnitude and $1^{\circ}$ for voltage angle \citep{IMSys_ref}, as an illustrative reference rather than an operational threshold (Section~\ref{sec:methods_baselines}). 

\begin{figure*}[t]
  \centering
  \includegraphics[width=0.75\linewidth]{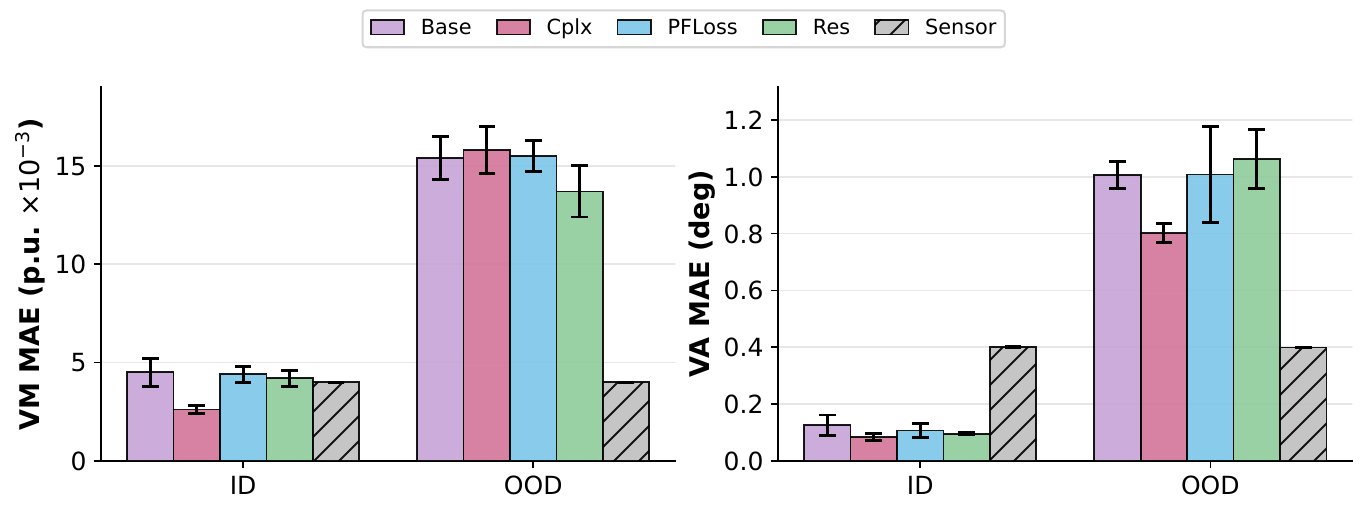}
  \caption{Summary of prediction error ($\text{MAE}^{Median}$) of physics-informed GNN variants in ID and OOD experiments ($n=5$).}
  \label{fig:performance_summary_gnn_models}
\end{figure*}

\subsubsection{Topology-Centric Evaluation}

Because our primary focus is assessing OOD generalization across diverse feeders, we employ a topology-centric evaluation. If global metrics were computed simply across all individual nodes in a heterogeneous test set, a large network (e.g., a 129-bus grid) would disproportionately skew the final calculation compared to a smaller network (e.g., a 15-bus grid). To ensure every test system is weighted equally, we define a single, representative graph-level error, denoted as $\varepsilon_G$, for each target variable. Let $\mathcal{G}$ denote the set of all test graphs, and $\mathcal{G}_j$ represent the set of test graphs for a specific evaluation scenario $j$. For a given graph $G \in \mathcal{G}$ with $N_G$ non-slack nodes,
the graph-level error is the Mean Absolute Error (MAE) across all nodes, defined as: $$\varepsilon_G = \text{MAE}_G = \frac{1}{N_G} \sum_{i=1}^{N_G} E_i$$

This approach provides a pure, unskewed measure of expected nodal accuracy per grid, answering the critical deployment question: \emph{If a model is deployed on a randomly selected grid, what is the typical expected error for that system?}

\subsubsection{Global Performance Metrics}

For final determination of model performance, we evaluate over a mixed dataset of networks. However, when testing on highly heterogeneous grids, error distributions are frequently skewed by isolated, extreme outliers. To provide a non-parametric measure of typical predictive capabilities, that resists outlier distortion while still providing a comprehensive view of model robustness, we aggregate the graph-level errors ($\varepsilon_G$) using the median, 25th percentile (Q1), and 75th percentile (Q3). For a given test scenario $j$, these are denoted as $\text{MAE}_{\mathcal{G}_j}^{Median}$, $\text{MAE}_{\mathcal{G}_j}^{Q1}$, and $\text{MAE}_{\mathcal{G}_j}^{Q3}$, respectively. Table \ref{tab:pinn_overview} shows that worst-case metrics generally appear  in the optimization literature, where constraint violation is the quantity of interest, rather than in typical power flow studies.
To characterize tail case behaviour (a necessity for safety-critical power systems) we extract the absolute maximum nodal error across the entire scenario,
indicating the empirical maximum test error across all graphs: $$\text{MaxAE}_{\mathcal{G}_j} = \max_{G \in \mathcal{G}_j} \left( \max_{i \in N_G} E_i \right)$$

By isolating typical expected accuracy from the observed maximum error, this metric framework provides a highly transparent and operationally relevant assessment of a solver's true reliability. The aggregation of these metrics across feeders and runs is detailed in Appendix~\ref{app:aggregation}.
\section{Results}\label{sec:results}

\subsection{Standard Performance Evaluation}\label{sec:standard_performance}

To evaluate the efficacy of the presented inductive biases, we first analyze the typical predictive performance across both in-distribution and out-of-distribution scenarios. Figure~\ref{fig:performance_summary_gnn_models} presents a summary of the median graph-level errors ($\text{MAE}^{Median}$) for each model variant.
Under known grid configurations (ID), the Complex-Valued (\texttt{Cplx}) model emerges as the strongest overall performer, yielding the lowest errors for both state variables ($0.0026$ p.u. and $0.0837^{\circ}$, respectively). We observe that all tested models fall below the illustrative meter tolerance presented in Section~\ref{sec:methods_baselines} in the median case, achieving highly accurate predictions.

The transition to unseen feeders (OOD) reveals a significant generalization bottleneck. 
Across all models and metrics, performance degrades by factors of three to eleven, highlighting the inherent difficulty neural networks face when generalizing to novel distribution grid structures. 
Furthermore, the OOD results demonstrate that no single inductive bias universally dominates. 
Instead, the models exhibit distinct specializations: though quite similar overall, the \texttt{Res} model achieves the lowest median error for voltage magnitude ($0.0137$ p.u.), while the \texttt{Cplx} model provides the most accurate voltage angle predictions ($0.8023^{\circ}$).

However, the intervals caused by multiple seed runs in Figure~\ref{fig:performance_summary_gnn_models} produce a performance overlap between model variants in many cases. To separate true improvement from variation induced by re-training, Table~\ref{tab:paired_seed_differences} reports the per-seed paired difference between each variant and the \texttt{Base} model. Of the twelve comparisons of voltage error, only two yield a confidence interval excluding zero, meaning that the mean paired difference over the five sampled runs is distinguishable from zero. The \texttt{Cplx} model improves in-distribution voltage magnitude by $0.0019$~p.u.\ $[0.0008, 0.0030]$ and out-of-distribution voltage angle by $0.2037^{\circ}$ $[0.1085, 0.2989]$. Every comparison involving the physics-constrained loss or the residual reformulation is statistically indistinguishable from the baseline in both regimes; the closest is the residual model's out-of-distribution voltage magnitude error ($0.0017$~p.u.\ $[-0.0001, 0.0035]$), whose interval only marginally includes zero. The benefit of the complex-valued design is therefore real but regime-dependent, appearing for magnitude under known feeders and for angle under unseen ones, while the remaining two biases yield no improvement that survives repeated training.

Furthermore, on unseen grids no model attains a median voltage magnitude error within the illustrative meter tolerance of $0.01$ p.u.
Only the \texttt{Cplx} model remains beneath the $1^\circ$ reference for voltage angle ($0.8023^{\circ}$), and it is also the only variant whose angle improvement over the baseline is statistically distinguishable.
Since this concerns the median case, these results do not represent the full picture of prediction accuracy for the evaluated models. 
To provide a more comprehensive analysis of these limitations, Section \ref{res:robustness_case_studies} presents case-centric studies isolating specific model behaviours and failure modes, with full benchmarking results provided in Appendix~\ref{app:complete_benchmarking_results}.

\begin{table*}[t]
\centering
\caption{Paired per-seed differences relative to the Base model (Base $-$ variant); positive values favour the variant. Intervals are two-sided 95\% $t$-intervals on the mean paired difference ($n=5$), with the seed as the experimental unit and approximately normal differences assumed. The comparisons are exploratory and not adjusted for multiplicity. The upper block reports voltage prediction error (Section~\ref{sec:standard_performance}) and the lower block the normalized power mismatch (Section~\ref{res:physical_consistency}).
Bold indicates an interval excluding zero.}
\label{tab:paired_seed_differences}
\setlength{\tabcolsep}{4pt}
\begin{tabular}{lcccc}
\toprule
\textbf{Model} & \multicolumn{2}{c}{\textbf{ID}} & \multicolumn{2}{c}{\textbf{OOD}} \\
 & \textbf{VM$_{\text{(p.u.)}}$} & \textbf{VA$_{\text{(deg)}}$} & \textbf{VM$_{\text{(p.u.)}}$} & \textbf{VA$_{\text{(deg)}}$} \\
\midrule
Cplx & \textbf{0.0019} {\scriptsize [0.0008, 0.0030]} & 0.0417 {\scriptsize [-0.0141, 0.0976]} & -0.0004 {\scriptsize [-0.0026, 0.0018]} & \textbf{0.2037} {\scriptsize [0.1085, 0.2989]} \\
PFLoss & 0.0001 {\scriptsize [-0.0007, 0.0008]} & 0.0193 {\scriptsize [-0.0158, 0.0543]} &  -0.0001 {\scriptsize [-0.0023, 0.0021]} & -0.0017 {\scriptsize [-0.2378, 0.2345]} \\
Res & 0.0004 {\scriptsize [-0.0010, 0.0017]} & 0.0305 {\scriptsize [-0.0188, 0.0798]} & 0.0017 {\scriptsize [-0.0001, 0.0035]} & -0.0557 {\scriptsize [-0.2534, 0.1420]} \\
\midrule
 & \textbf{$\bm{S}_{\text{nodal}}$} & \textbf{$\bm{S}_{\text{slack}}$} & \textbf{$\bm{S}_{\text{nodal}}$} & \textbf{$\bm{S}_{\text{slack}}$} \\
\midrule
Cplx & \textbf{0.6449} {\scriptsize [0.3614, 0.9284]} & 0.0678 {\scriptsize [-0.0226, 0.1581]} & -0.3544 {\scriptsize [-3.7008, 2.9920]} & 0.3067 {\scriptsize [-0.7822, 1.3956]} \\
PFLoss & 0.1392 {\scriptsize [-0.0591, 0.3375]} & 0.0462 {\scriptsize [-0.0263, 0.1188]} & 1.1717 {\scriptsize [-2.7207, 5.0642]} & 0.3031 {\scriptsize [-0.5900, 1.1962]} \\
Res & 0.1835 {\scriptsize [-0.0159, 0.3828]} & 0.0399 {\scriptsize [-0.0139, 0.0937]} & 1.2719 {\scriptsize [-0.9564, 3.5002]} & 0.3622 {\scriptsize [-0.2087, 0.9331]} \\
\bottomrule
\end{tabular}
\end{table*}

\subsection{Physical Consistency}\label{res:physical_consistency}

To verify whether \texttt{PFLoss} achieves its learned objective of preserving the power balance equations, we benchmark physical consistency using two metrics. Standard literature typically reports nodal apparent power mismatch \citep{rivera2025pf}, $S_{\text{nodal}}$, which we include for comparability. In distribution networks, however, this quantity is highly sensitive to errors in the voltage differences between adjacent buses. Feeders consist of short, low-impedance lines, so a small error in the predicted voltage difference across a branch implies a large error in its current through $I = \Delta V / Z$, and hence in the nodal power balance. As a complementary view better suited to these grids, we therefore also isolate the apparent power mismatch at the slack bus alone, $S_{\text{slack}}$. In distribution networks the slack bus represents the primary substation transformer, whose considerably higher leakage reactance reduces this sensitivity, and the net power drawn from the upstream grid is the quantity of most direct operational interest for capacity planning and thermal limit monitoring. For evaluation, both metrics are normalized by the grid's maximum gross throughput, $S_{\text{gross}}$, to remove the scale of feeder loading. Full definitions are given in Appendix~\ref{app:consistency}.

\begin{figure}[t]
  \centering
  \includegraphics[width=\linewidth]{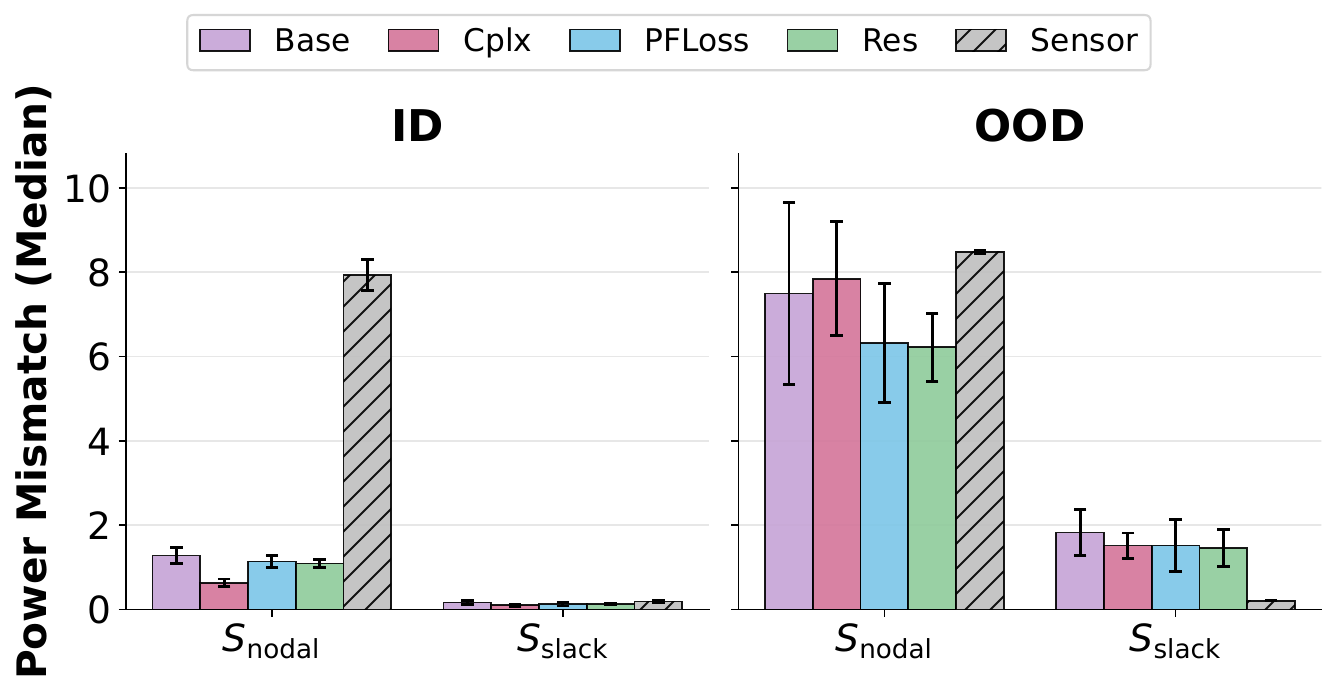}
  \caption{Evaluation of physical consistency using median power mismatch metrics for $S_{\text{nodal}}$ and $S_{\text{slack}}$ ($n=5$).}
  \label{fig:physical_consistency_comparison}
\end{figure}

Figure~\ref{fig:physical_consistency_comparison} reports median results for both metrics for all variants under ID and OOD conditions, alongside a simulated smart meter. \texttt{PFLoss} attains a lower median mismatch than \texttt{Base} in all four settings, but none of these reductions is distinguishable from zero in the paired analysis (Table~\ref{tab:paired_seed_differences}), and in none of them is it the strongest variant. \texttt{Cplx} attains the lowest in-distribution mismatch under both metrics, and \texttt{Res} the lowest out-of-distribution.
The physics-constrained loss therefore yields no improvement in physical consistency that survives repeated training, including on nodal mismatch, which is closely related to the quantity its $L_{\text{phys}}$ term minimizes. The simulated sensor results illustrate how strongly physical consistency evaluation depends on the choice of formulation. Note the scale difference between the two metrics. A measurement perturbed only within standard hardware tolerance yields a median nodal mismatch of $7.9341$ p.u.\ in-distribution, roughly six times that of any learned model. This reflects the independent perturbation of adjacent buses, which produces errors in their voltage differences of the same order as the perturbations themselves. Under $S_{\text{slack}}$ the sensor is comparable to the models in-distribution and remains stable out-of-distribution where they degrade, as expected of a measurement that perturbs the true state rather than predicting it, and is thus unaffected by training.

We present these results as a diagnostic for the physics loss rather than as a basis for ranking models. Neither metric has an established operational threshold for distribution grid control, and the nodal formulation is dominated by small errors in the voltage differences between adjacent buses. Establishing an operationally meaningful consistency metric for distribution networks, together with defensible thresholds and a dedicated analysis of failure behaviour, is a study in its own right, which we identify as future work in Section~\ref{sec:future}. The remainder of our results accordingly focus on voltage prediction accuracy.

\subsection{Analytical Baselines}

\begin{figure}[t]
  \centering
  \includegraphics[width=\linewidth]{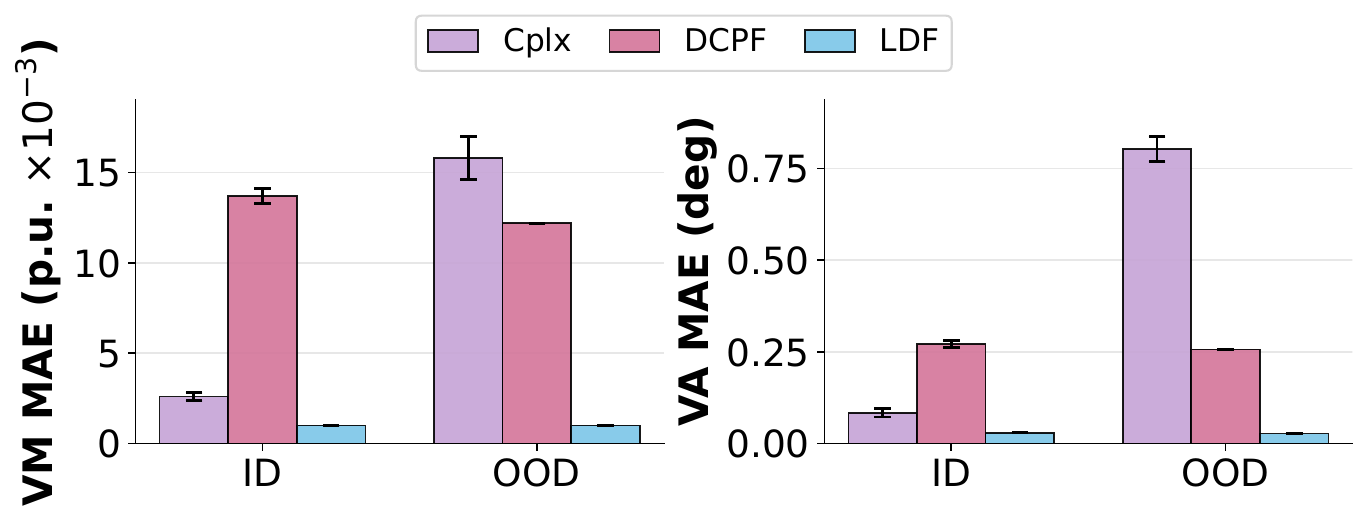}
  \caption{Performance summary ($\text{MAE}^{Median}$) of \texttt{Cplx} model against the linearized analytical baselines ($n=5$).}
  \label{fig:performance_summary_analytic_baselines}
\end{figure}

To better understand the operational utility of using these physics-informed GNNs, it is crucial to investigate how their performance compares relative to established power flow approximation techniques. 
To achieve this, we benchmark our best-performing model, \texttt{Cplx}, against the DCPF and LDF baselines from Section~\ref{sec:methods_baselines}.
As shown in Figure \ref{fig:performance_summary_analytic_baselines}, the \texttt{Cplx} model outperforms DCPF within the ID scenario, demonstrating that on familiar topologies the neural architecture captures the non-linear behaviour of the grid better than this common simplified solver. It does not, however, reach the accuracy of LDF, which achieves lower errors than every neural variant in both settings ($0.0010$ p.u.\ and $0.0292^{\circ}$ in-distribution, against $0.0026$ p.u.\ and $0.0837^{\circ}$ for \texttt{Cplx}). The accuracy of LDF reflects the small voltage deviations of most operating points in the dataset, under which its linearization error is small. In the OOD setting, the gap widens to more than an order of magnitude ($0.0158$ versus $0.0010$ p.u.\ and $0.8023^{\circ}$ versus $0.0272^{\circ}$), and \texttt{Cplx} is also outperformed by DCPF on both state variables.

Prior benchmarking studies, which compared against DCPF, have found GNN solvers to be outperformed on unseen topologies \citep{hansen_power_2022, yaniv_towards_2023, okoyomon2025framework}, which our OOD results confirm. In this distribution setting, however, the distribution-specific LDF outperforms every neural variant in-distribution as well. That a neural architecture trained explicitly on full AC power flow data is outperformed on unseen systems even by DC power flow, whose voltage-magnitude ``prediction'' is simply a flat profile of $1.0$ p.u.\ at every bus, ignoring line resistance and reactive power entirely,
highlights the severity of the cross-system generalization gap. This result underscores that while physics-informed biases can improve ID performance and stabilize the training process, truly generalizable machine learning estimators must still overcome significant error margins before they can reliably augment or replace traditional analytical solvers in arbitrary, unseen power networks. To verify that these findings are not an artifact of an unrepresentative neural architecture, we additionally benchmark our variants against two learned baselines from the recent literature in Appendix~\ref{app:learned_baselines}.

\subsection{Model Robustness}\label{res:robustness_case_studies}

Most studies proposing ML-based power flow solvers present only summarized statistics and average-case results for model evaluation. However, due to the critical reliability requirements of grid infrastructure, relying solely on medians or aggregated error metrics of RMSE and MAE can paint an incomplete and potentially misleading picture of model readiness. To truly understand the accuracy, robustness, and pitfalls of these models, a comprehensive analysis of specific test scenarios and failure modes is required.

Table \ref{tab:robustness_case_studies} presents more comprehensive, detailed metrics for two distinct case studies drawn from our experimentation suite. We evaluate the in-distribution baseline alongside the most difficult OOD scenario across the test scenarios, representing a specific scale-generalization challenge. 
To quantify model robustness, we report the interquartile range (IQR), defined as $[Q1, Q3]$, while the absolute maximum error approximates the worst-case performance for each model.

\begin{table*}[htbp]
\centering
\caption{Case studies for evaluation of model robustness ($n=5$). MaxAE is the mean of the five per-run maxima, and $\pm$ denotes the standard deviation across runs.}
\label{tab:robustness_case_studies}
\begin{tabular}{llcccc}
\toprule
\textbf{Testing Grid} & \textbf{Model} & \multicolumn{2}{c}{\textbf{VM (p.u.)}} & \multicolumn{2}{c}{\textbf{VA (deg)}} \\
 &  & \textit{MAE$^{\text{Median}}$} \tiny{[Q1, Q3]} & \textit{MaxAE} & \textit{MAE$^{\text{Median}}$} \tiny{[Q1, Q3]} & \textit{MaxAE} \\
\midrule
ID (All) & Base & 0.0045 {\tiny $\pm$ 0.0007} {\scriptsize [0.0032, 0.0064]} & 0.0409 {\tiny $\pm$ 0.0052} & 0.1254 {\tiny $\pm$ 0.0357} {\scriptsize [0.0859, 0.1819]} & \textbf{1.6814} {\tiny $\pm$ 0.4069} \\
ID (All) & Cplx & \textbf{0.0026 {\tiny $\pm$ 0.0002}} {\scriptsize [\textbf{0.0018}, \textbf{0.0045}]} & \textbf{0.0291} {\tiny $\pm$ 0.0051} & \textbf{0.0837 {\tiny $\pm$ 0.0114}} {\scriptsize [\textbf{0.0618}, \textbf{0.1224}]} & 1.7508 {\tiny $\pm$ 0.1702} \\
ID (All) & PFLoss & 0.0044 {\tiny $\pm$ 0.0004} {\scriptsize [0.0032, 0.0062]} & 0.0531 {\tiny $\pm$ 0.0323} & 0.1062 {\tiny $\pm$ 0.0244} {\scriptsize [0.0771, 0.1513]} & 2.1320 {\tiny $\pm$ 1.4375} \\
ID (All) & Res & 0.0042 {\tiny $\pm$ 0.0004} {\scriptsize [0.0031, 0.0057]} & 0.0384 {\tiny $\pm$ 0.0036} & 0.0949 {\tiny $\pm$ 0.0036} {\scriptsize [0.0722, 0.1375]} & 2.8113 {\tiny $\pm$ 2.5069} \\
\midrule
LV-rural1 & Base & 0.0470 {\tiny $\pm$ 0.0271} {\scriptsize [0.0460, 0.0479]} & 0.1249 {\tiny $\pm$ 0.0761} & 7.2003 {\tiny $\pm$ 7.7054} {\scriptsize [7.1770, 7.2240]} & 21.0029 {\tiny $\pm$ 25.0517} \\
LV-rural1 & Cplx & 0.1242 {\tiny $\pm$ 0.0628} {\scriptsize [0.1235, 0.1249]} & 0.3742 {\tiny $\pm$ 0.2538} & 4.1821 {\tiny $\pm$ 1.0123} {\scriptsize [4.1512, 4.2139]} & 12.1068 {\tiny $\pm$ 3.3234} \\
LV-rural1 & PFLoss & 0.0549 {\tiny $\pm$ 0.0263} {\scriptsize [0.0544, 0.0555]} & 0.1459 {\tiny $\pm$ 0.0517} & \textbf{3.7216 {\tiny $\pm$ 1.6252}} {\scriptsize [\textbf{3.6991}, \textbf{3.7439}]} & \textbf{9.5063} {\tiny $\pm$ 3.3810} \\
LV-rural1 & Res & \textbf{0.0210 {\tiny $\pm$ 0.0089}} {\scriptsize [\textbf{0.0202}, \textbf{0.0219}]} & \textbf{0.0632} {\tiny $\pm$ 0.0211} & 4.2079 {\tiny $\pm$ 1.2103} {\scriptsize [4.1781, 4.2385]} & 14.5505 {\tiny $\pm$ 9.3872} \\
\bottomrule
\end{tabular}
\end{table*}

\subsubsection{The Illusion of Safety}

The ID scenario augments typical research settings by evaluating models on a mixed collection of heterogeneous networks, all drawn from the training distribution. In this setting, the \texttt{Cplx} model is the strongest performer. It achieves best-in-class results across median nodal prediction accuracy, confirming its superior capacity to capture the intricacies of known, heterogeneous distribution grids. However, a critical nuance emerges when comparing the IQR tightness to the maximum-error predictions. In the ID setting, the IQR widths are exceptionally narrow (for example, the \texttt{Base} model's VM IQR width is $\approx 0.0032$ p.u.). Yet, the MaxAE values are roughly $9$ to $30\times$ the median values and all exceed the sensor tolerances of Section~\ref{sec:metrics}, showing that models capturing typical in-distribution behaviour can still produce severe localized errors, although these maxima do not establish how often such errors occur.

Particularly concerning is the behaviour of the \texttt{PFLoss} model, which records the highest maximum voltage magnitude error ($0.0531$ p.u.) and quite poor maximum voltage angle error ($2.1320^{\circ}$) despite competitive median accuracy. In contrast, the plain \texttt{Base} model records lower maximum error results for both of these, including the lowest maximum angle error of the four variants ($1.6814^{\circ}$). Physics-constrained training therefore does not inherently bound tail-end errors. For practitioners this is a crucial reliability caveat: deploying a physics-informed loss function does not guarantee ``physics-safe'' operation across all localized nodes. Standard aggregate metrics place all models well below the illustrative meter tolerances, while this failure-mode analysis exposes severe localized errors.

\subsubsection{Cross-System Generalization Failure}

The LV-rural1 test case has only 15 buses and is the smallest grid evaluated.
Analysis from a previous study demonstrates that this network has very different degree distribution and eigenvector geometry from all other training graph types in the benchmarking suite (Appendix~\ref{app:topology-ood}), making this the most challenging OOD test case.
Furthermore, GNNs trained on larger power networks have also been shown to struggle particularly when predicting exceptionally small grids \cite{jeddi_physics-informed_2021}, making this a perfect test for cross-system generalization.

All variants achieve their worst median performance on this test case, and no architecture is uniformly better. The \texttt{Res} model attains the lowest magnitude error ($0.0210$ p.u.) and the \texttt{PFLoss} model the lowest angle error ($3.7216^{\circ}$), while the \texttt{Cplx} model degrades sharply in magnitude ($0.1242$ p.u., roughly $48$ times its error on the in-distribution scenario) and the \texttt{Base} model produces the largest angle MaxAE in this study ($21.0029^{\circ}$). We hypothesize that the complex phase relationships learned from larger, meshed MV networks translate poorly to the localized, strictly radial geometry of this small grid, although Appendix~\ref{app:topology-ood} establishes only the structural dissimilarity of this feeder rather than the cause of its errors. The inductive biases more generally offer no consistent protection: each is best on one state
variable and mediocre or worse on the other.

The most telling feature of these results, however, is the structure of their variability. One would expect a model encountering an unfamiliar network to produce erratic, visibly unstable predictions. Instead the interquartile ranges here are comparable to or tighter than those in the in-distribution setting, despite being very wrong.
Notably, across independent initializations the same model can vary by a standard deviation even larger than its own mean error and orders of magnitude wider than its within-run spread, such as in the case of the \texttt{Base} model. 
Such failures can be more dangerous than widely dispersed errors, since a narrow spread of errors across graphs creates a false impression of stability, although this dispersion is not a measure of predictive uncertainty.

\subsubsection{Most Severe Errors}

Examining the MaxAE values across all eleven scenarios (Table~\ref{tab:model_performance_raw}) reveals a pattern that median metrics conceal. The three largest voltage magnitude MaxAE values in the study all belong to the \texttt{Cplx} model: $0.6428$ p.u.\ on LV-rural3, $0.3742$ p.u.\ on LV-rural1, and $0.2531$ p.u.\ on MV-semiurb. The largest MaxAE of any other variant is $0.1729$ p.u. LV-rural1 and LV-rural3 are also the two scenarios on which \texttt{Cplx} records its worst median magnitude error, indicating a degradation on these feeders that is not confined to their extreme values. The variant achieving the best median accuracy in-distribution thus also produces the most extreme localized failures out-of-distribution, which is precisely the divergence between average-case and worst-case behaviour that motivates this evaluation framework. The \texttt{Res} model exhibits a comparable pattern for voltage angle, recording the largest MaxAE of all variants on four of the eleven scenarios despite competitive median performance.

For reference, the analytical baselines exhibit markedly smaller extremes: their largest MaxAE values in the study are $0.0577$~p.u.\ and $2.30^{\circ}$, against $0.6428$~p.u.\ and $21.00^{\circ}$ for the neural variants, and their angle MaxAE is lower than that of every neural variant on all ten out-of-distribution scenarios. The tail behaviour reported here is therefore primarily a property of the learned models specifically rather than power flow approximation in general.
\section{Discussion}

\begin{table}[t]
\centering
\caption{Training cost and model capacity. Training time is averaged across all experimental scenarios and seeds ($\pm$ standard deviation across these 55 runs); capacity is architecture-determined and does not vary by feeder.}
\label{tab:model_efficiency_training_capacity}
\begin{tabular}{@{}lcc@{}}
\toprule
\textbf{Model} & \textbf{Capacity} & \textbf{Training time (s)} \\
\midrule
Base   & $302.7$K & 1615.9 {\scriptsize $\pm$ 58.5} \\
Cplx   & $603.4$K & 3504.2 {\scriptsize $\pm$ 326.6} \\
PFLoss & $302.7$K & 3033.6 {\scriptsize $\pm$ 271.6} \\
Res    & $302.7$K & 1611.8 {\scriptsize $\pm$ 62.3} \\
\bottomrule
\end{tabular}
\end{table}

\subsection{Model Efficiency}\label{method:model-efficiency}

\begin{figure*}[t]
  \centering
  \includegraphics[width=0.8\linewidth]{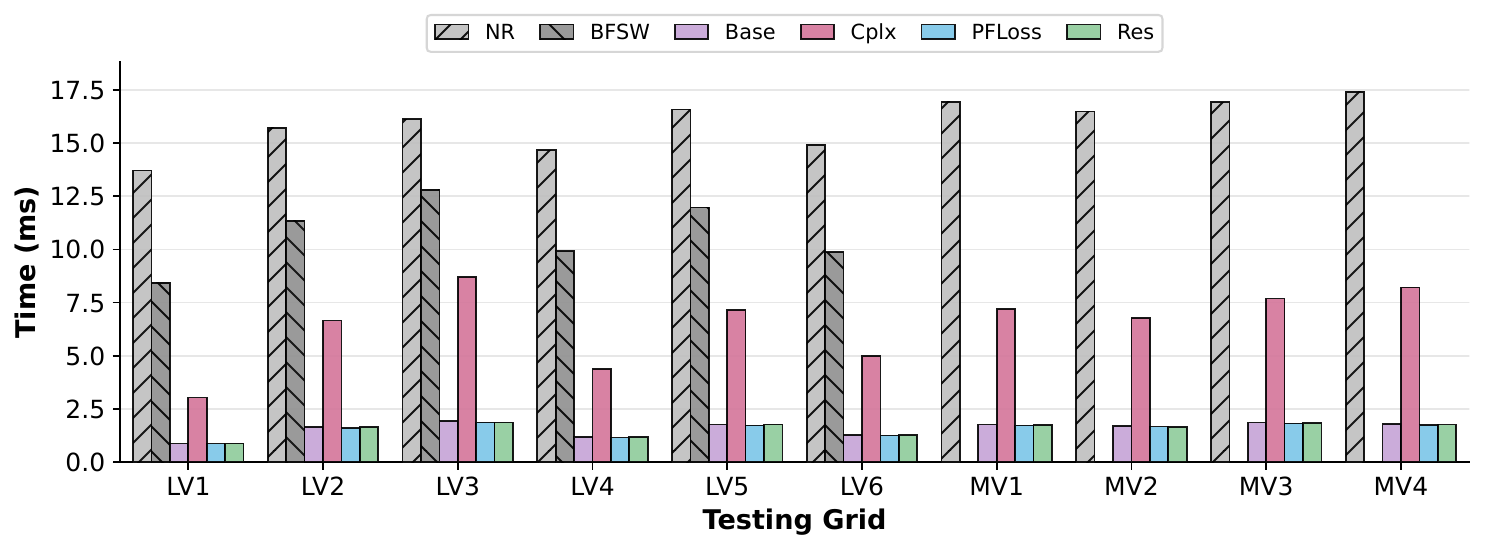}
  \caption{Per-network inference latency across the ten test feeders in ms. BFSW only converges for LV networks.}
  \label{fig:inference_time}
\end{figure*}

While approximation accuracy is important for safety-critical systems, deploying these models in operational environments also requires a careful evaluation of computational efficiency. Table~\ref{tab:model_efficiency_training_capacity} summarizes training cost and model capacity, while Figure~\ref{fig:inference_time} reports per-network inference latency against two high-fidelity analytical solvers. Training was conducted on a workstation with an NVIDIA RTX 5000 Ada GPU with 32 GB of GDDR6 VRAM.
Training time was averaged across all scenarios and seeds for simplicity. To ensure a fair comparison to the analytical models, inference timings for all models were measured on identical hardware (Apple M3, 16\,GB RAM, CPU only) and averaged over 300 sequential runs per feeder.

The \texttt{Res} architecture presents the most lightweight inductive bias. It incurs virtually no additional computational overhead across any of the tracked efficiency metrics, making it an easily integrated structural addition that does not rely on expanded hardware resources. Conversely, the \texttt{PFLoss} model introduces a significant computational burden during the training phase ($1615.9$ $\shortrightarrow$ $3033.6$ seconds), requiring approximately double the training time of the \texttt{Base} model. This increase is driven entirely by the cost of backpropagating through the complex, highly non-linear AC power balance equations during the fine-tuning stage. Notably, this penalty is strictly confined to training; its inference time and model capacity remain identical to the baseline. For practical applications, this training overhead could be mitigated by utilizing linearized power flow equations within the loss function or by dynamically reducing the number of fine-tuning epochs once physical feasibility thresholds are met.

The most significant efficiency trade-offs are observed with the \texttt{Cplx} model. It incurs the highest computational burden because it natively represents complex values as pairs of real numbers throughout its layers and activations. Compared to the \texttt{Base} model, it requires approximately $2\times$ the training time, $3$--$4\times$ the inference latency, and $2\times$ the parameter capacity.

\subsubsection{Inference Time}

To situate these latencies operationally, we benchmark all variants against Newton-Raphson (NR) and backward/forward sweep (BFSW) on the same hardware and with a convergence tolerance of $10^{-8}$~MVA. We use pandapower~\citep{thurner2018pandapower} (version 2.14.10, with numba acceleration) as the reference implementation, as it is a widely used open-source tool and was used to generate the ENGAGE dataset. NR fails to converge from a flat start ($1.0$~p.u., $0^{\circ}$) on \emph{every} feeder in the dataset, irrespective of the iteration limit, and is therefore initialized with a DC power flow. Furthermore, BFSW fails to converge on all four MV feeders regardless of initialization, returning singular-matrix errors, and is therefore reported for the six radial LV feeders only.

The NR and BFSW timings cover a complete call to pandapower, which in addition to the iterative solution includes network conversion and admittance-matrix construction. The neural timings cover a forward pass which performs feature preparation, missing-value handling, and then prediction on a graph already in tensor form. Both are measured after a warm-up run. The two timings thus do not cover identical stages, and since part of each pandapower call is overhead from creating the PYPOWER case, the timings likely favour the neural models.

For the configurations that did converge, NR requires $13.70$--$17.41$~ms and BFSW $8.43$--$12.80$~ms per network, against $0.88$--$1.94$~ms for the \texttt{Base} architecture and $3.04$--$8.71$~ms for \texttt{Cplx}. For reference, the DCPF formulations run in $3.29$--$6.20$~ms. 
In this single-network workflow, the \texttt{Base} model is approximately $9.6\times$ faster than NR in the median case and $7.4\times$ faster than BFSW on the radial feeders, while the more expensive \texttt{Cplx} model retains a median advantage of $2.4\times$ over NR.

These results do not reproduce the orders-of-magnitude accelerations frequently reported for neural power flow surrogates. We attribute this primarily to scale, since typical European distribution feeders are considerably smaller and more sparsely connected than the large transmission systems on which such speedups are usually demonstrated, leaving little numerical work for a neural model to displace. Larger accelerations in this setting are more likely to come from batched inference. A neural surrogate evaluates a batch of networks as a single block-diagonal graph in one forward pass, whereas a heterogeneous collection of feeders presents a distinct admittance sparsity pattern per case, so no LU factorization can be shared in batched NR and every network pays its full factorization cost. This distinction is most consequential precisely where many related cases must be solved at once, as in capacity planning studies and N-$k$ contingency screening. We do not benchmark this regime here, as a meaningful batched comparison would require solver implementations engineered for batch execution, since standard multiprocessing would introduce excess framework overheads. We therefore identify batched inference as a promising direction rather than a demonstrated result.

\subsubsection{Model Capacity Ablation}

To assess whether the \texttt{Cplx} model's advantage stems from its structural alignment with grid physics rather than from its parameter count, we construct a capacity-matched real-valued control (\texttt{Base-L}) by scaling the hidden dimension of the message-passing layers from 128 to 192, yielding $624.2$K parameters against the \texttt{Cplx} model's $603.4$K. Both are evaluated under the ID and OOD protocols.

\begin{figure}[t]
  \centering
  \includegraphics[width=\linewidth]{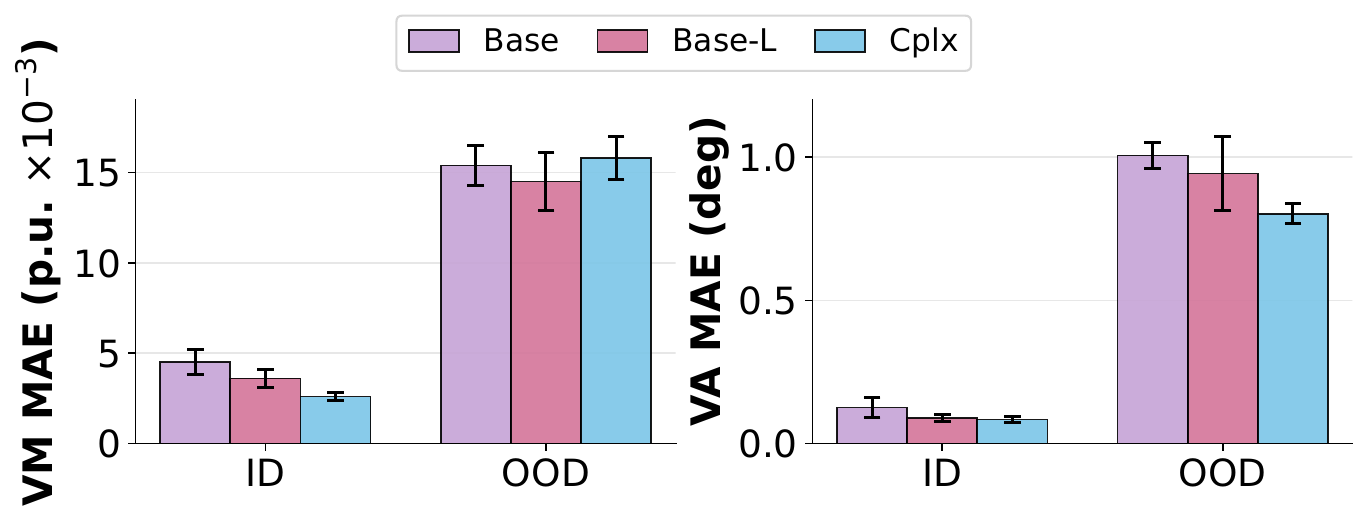}
  \caption{\texttt{Cplx} model compared to normal and capacity-matched \texttt{Base} models in ID and OOD experiments ($n=5$).}
  \label{fig:model_capacity_experiment}
\end{figure}

Figure~\ref{fig:model_capacity_experiment} shows that additional capacity is beneficial in its own right since \texttt{Base-L} improves on \texttt{Base} for all four metrics. The \texttt{Cplx} model nonetheless retains its advantage where it had one previously. It achieves lower ID voltage magnitude error than \texttt{Base-L} ($0.0026$ vs $0.0036$ p.u.) and lower OOD voltage angle error ($0.8023^{\circ}$ vs $0.9426^{\circ}$), which are the same two metrics on which it was distinguishable from \texttt{Base} in Section~\ref{sec:standard_performance}. For ID voltage angle the two models are close ($0.0837^{\circ}$ vs $0.0900^{\circ}$), and for OOD voltage magnitude \texttt{Base-L} is the stronger model ($0.0145$ vs $0.0158$ p.u.), a metric on which \texttt{Cplx} likewise did not improve on the plain \texttt{Base} model. This indicates that the observed improvements are not explained by parameter count alone, but suggests that the complex-valued design assists in capturing the physical coupling of the voltage variables. Since \texttt{Base-L} retains the real-valued output and loss, this ablation controls for capacity but not for the phasor output and loss that accompany the complex representation.

\subsection{Real-World Application}

The results of this study inform on the operational readiness of the evaluated physics-informed models in distribution grids. In in-distribution scenarios, they infer nodal voltages from power injections with median errors within the illustrative meter tolerances, most accurately when using the \texttt{Cplx} architecture, although LDF achieves lower errors on the same feeders. However, median accuracy alone is a weak basis for deployment in power systems, where the consequence of a prediction is determined by its worst localized error rather than its typical one. Our out-of-distribution and failure-case analyses show that the evaluated models exhibit substantial cross-feeder errors, including severe localized errors that median metrics conceal. Whether such errors are acceptable depends on the application, the domain in which a model has been validated, and the safeguards and fallback arrangements in place, none of which is evaluated here. Our results therefore indicate that these models require application-specific validation and safeguards before operational deployment, particularly on feeders unlike those seen in training.

Thus, our results do not support recommending any individual inductive bias on the basis of predictive performance.
Where a practitioner must nonetheless choose, the decision is better made on computational grounds than on expected accuracy. Residual learning is effectively free, adding no overhead in training or inference. The physics-constrained loss roughly doubles training time while leaving inference unchanged, but neither its predictive accuracy nor its physical consistency improves on the \texttt{Base} model by a margin distinguishable from zero in the paired analysis. The complex-valued architecture is the most accurate in-distribution but the most expensive and produces the three largest magnitude MaxAE values in the OOD portion of the study. In all cases, deployment should be preceded by validation on structurally similar feeders, as none of the biases evaluated here provides protection against the cross-system failures reported in Section~\ref{res:robustness_case_studies}.

\subsection{Evolving Evaluation Methodologies}

A central focus of this study is demonstrating the wide accuracy range and failure modes of neural network models for distribution grid voltage prediction. While foundational benchmarking datasets and frameworks, such as those presented in \citep{varbella_powergraph_2024}, have made efforts in improving comparability of ML models in this fast-moving domain, our results highlight that the transition to deployment-ready models requires evolving these protocols away from average case evaluations (i.e. their presented global MSE metric) and towards comprehensive assessments across several heterogeneous scenarios. A notable exception is the work from \citet{rivera2025pf}, who earnestly attempt to include a more comprehensive scope in their benchmarking framework via mean and maximum metrics. However, they too ultimately rely solely on average-case results when presenting model comparisons. As our OOD and maximum empirical error analyses demonstrate, relying on means and medians obscures the systematic biases and extreme outliers that define a model's true reliability, underscoring the need for generalization studies that may reveal critical operational deficiencies. The research community must ubiquitously adopt failure-case analyses and maximum boundary metrics to ensure transparency in model capabilities. For distribution networks in particular, our results indicate that a distribution-specific linearization such as LDF provides a considerably more demanding reference than DCPF, and should accompany evaluations of neural power flow solvers in this setting.

\subsection{Future Work}\label{sec:future}

Our conclusions are scoped to the setting evaluated here. All results concern message-passing GNNs using GraphConv and thus do not extend definitively to physics-informed neural solvers as a class. The ENGAGE dataset comprises ten synthetic, balanced single-phase-equivalent feeders derived from European networks, containing only PQ and slack buses, with fixed configurations and only converged operating points retained. The shared hyperparameters and physics-loss configuration were selected on a validation split containing samples of all ten feeders, so that held-out feeders were not excluded from model selection. Since such exposure would be expected to favour out-of-distribution performance, the reported generalization gap is, if anything, conservative. Expanding this evaluation framework to more techniques and to non-European, unbalanced, three-phase networks (such as Smart-DS \citep{krishnan2017smart}) will further test the applicability and limits of physics-informed methods in vastly different structural and operational environments, including whether the advantage of linearized baselines observed here persists in unbalanced networks, where they require detailed models of the coupling between phases. Similarly, the complex-valued design couples its internal representation with a phasor output and loss, and disentangling their individual contributions is left to future work.

Relatedly, our findings on operational readiness, including those concerning physical feasibility, should be read as demonstrations within this setting rather than as general claims about the suitability of neural solvers for real-world operation. A dedicated study of physical consistency in distribution networks is needed to establish an operationally meaningful mismatch metric with defensible thresholds and examine whether alternative formulations or weightings of the physics-constrained loss can improve feasibility where the soft penalty evaluated here does not.

To bridge the cross-feeder generalization gap, future research must also address the structural limitations of current message-passing frameworks. Investigating robust OOD detection mechanisms will be critical, allowing real-world systems to confidently flag unfamiliar networks and fall back to traditional solvers when uncertainty thresholds are exceeded.
Complementary to this, downstream formulations that 
treat surrogate prediction error as an additional source of distributional uncertainty offer probabilistic guarantees 
and may prove more tractable than bounding the predictor itself \citep{nawaz2026uncertainty}, while hard constraint projection methods or differentiable optimization layers 
could provide the mathematical guarantees required for closed-loop control \citep{nawaz2026physics}. 
\section{Conclusion}

This study systematically evaluates physics-informed Graph Neural Networks for voltage prediction in active distribution grids. By analyzing physics-constrained loss functions, complex-valued networks, and residual learning, we demonstrate that carefully chosen inductive biases provide distinct engineering trade-offs. Most notably, the Complex-Valued (\texttt{Cplx}) architecture attains the highest median accuracy among the evaluated variants on familiar feeders, indicating an increased ability
to capture the non-linear coupling of grid variables. 
However, we also demonstrate that standard average-case statistics paint an incomplete picture of model readiness. 
By shifting our evaluation from aggregated medians to empirical maximum test error and out-of-distribution evaluations, we expose severe limitations that standard benchmarks frequently obscure, highlighting the shortcomings of the evaluated neural power flow solvers. On unseen networks, the evaluated models exhibit substantial errors, so that application-specific validation and safeguards are required before operational deployment, particularly for tasks such as closed-loop control.

Moving forward, machine learning methodologies in power systems must shift their focus. Transitioning neural solvers from academic benchmarks to reliable, deployment-ready tools will require the community to prioritize rigorous cross-system generalization, worst-case boundary analysis, and the development of architectural paradigms that can strictly enforce operational constraints.

\printcredits

\bibliographystyle{cas-model2-names}

\bibliography{references}

\appendix
\begin{table*}[t!]
\centering
\caption{ENGAGE Dataset Statistics (averaged over samples)}
\label{tab:engage_stats}
\begin{tabular}{llccccccccc}
\toprule
\textbf{Grid} & \textbf{Type} & \textbf{Buses} & \textbf{Lines} & \textbf{Voltage} & \textbf{R/X} & \textbf{Total $\bm{|P_{\text{PQ}}|}$} & \multicolumn{3}{c}{\textbf{Line Loading}} \\
 & & & & \textbf{(kV)} & \textbf{Ratio} & \textbf{(MW)} & \textbf{Min (\%)} & \textbf{Max (\%)} & \textbf{Avg (\%)} \\
\midrule
\textbf{LV1} & Rural & 15 & 14 & 0.4 & 2.41 & 0.17 & 0.15 & 65.28 & 16.70 \\
\textbf{LV2} & Rural & 97 & 96 & 0.4 & 2.55 & 0.34 & 0.07 & 57.69 & 9.93 \\
\textbf{LV3} & Rural & 129 & 128 & 0.4 & 2.55 & 0.71 & 0.06 & 92.09 & 17.42 \\
\textbf{LV4} & Semi-urban & 44 & 43 & 0.4 & 2.52 & 0.34 & 0.73 & 99.99 & 29.98 \\
\textbf{LV5} & Semi-urban & 111 & 110 & 0.4 & 1.58 & 0.74 & 0.01 & 91.10 & 19.56 \\
\textbf{LV6} & Urban & 59 & 58 & 0.4 & 1.56 & 0.57 & 0.25 & 99.83 & 22.15 \\
\textbf{MV1} & Commercial & 106 & 113 & 20.0 & 1.92 & 36.01 & 0.01 & 99.30 & 13.84 \\
\textbf{MV2} & Rural & 97 & 102 & 20.0 & 2.62 & 22.80 & 0.08 & 48.46 & 10.21 \\
\textbf{MV3} & Semi-urban & 120 & 127 & 20.0 & 2.78 & 52.55 & 0.01 & 88.82 & 18.53 \\
\textbf{MV4} & Urban & 136 & 148 & 10.0 & 1.13 & 26.18 & 0.05 & 99.99 & 15.60 \\
\bottomrule
\end{tabular}
\end{table*}

\section{Experiment Setting} \label{app:experiment-setting}

\subsection{Dataset}

As depicted in Table \ref{tab:engage_stats}, the ENGAGE dataset is a heterogeneous distribution grid dataset, with 10 unique base distribution grids varying in size and type \citep{okoyomon_2025_15464235}. All grids are generated by applying Powerdata-gen \citep{powerdata-gen} to the low- and medium-voltage distribution networks from SimBench's scenario 1 (future grid with normal increase of distributed energy resources). Each grid type has 300 test cases with varying network loading conditions. Thus, our leave-one-out protocol tests scale generalization, as models trained on pools of smaller grids are tested on larger grids and vice versa. These distribution networks have fundamentally different characteristics than transmission networks, as they have much lower voltage levels, tend to be more radial in structure, and have much higher R/X ratios (approx. 0.1--0.3 is typical for transmission grids while distribution grids can have 1.0--5.0 or more, leading to large voltage drops).

\subsection{Preprocessing} \label{app:preprocessing}

Node inputs are the active and reactive power injections in MW and Mvar, numerically equal to per-unit values on the $1$~MVA system base, and the hop distance to the slack bus, to which the four global slack features are concatenated. Edge inputs are line resistance and reactance in p.u., and targets are voltage magnitude in p.u. and voltage angle in degrees. Differences in feature scale are absorbed by the batch normalization following each layer.

\subsection{Hyperparameter Search} \label{app:hyperparameters}

The \texttt{Base} configuration was selected by grid search, conducted in stages over subsets of the following space rather than as a single full factorial grid: number of message-passing layers $\{3, \dots, 10\}$, hidden dimension $\{64, 128, 256\}$, initial learning rate $\{10^{-2}, 10^{-3}, 10^{-4}, 10^{-5}\}$, batch size $\{64, 128, 256\}$, and epoch budget from $1000$ to $8000$ in steps of $100$. The selected configuration (seven layers, hidden dimension $128$, initial learning rate $10^{-4}$, batch size $128$, and $4000$ epochs) was applied unchanged to all variants.
This search, and the selection of the physics-loss configuration in Appendix~\ref{app:pf_loss_training_ablation}, were performed once on the ID validation split, which contains samples of all ten feeders, and the selected settings were applied unchanged to all variants and scenarios.

\subsection{Aggregation and Dispersion} \label{app:aggregation}

For each run, the scenario-level statistics of Section~\ref{sec:metrics} are computed over the graph-level errors $\varepsilon_G$ of that scenario, and $\text{MaxAE}$ over all nodes of all its graphs. The OOD summaries additionally take, for each run, the median of the ten feeder-level values, so that each held-out feeder is weighted equally irrespective of its size, consistent with the topology-centric evaluation. Unless stated otherwise, reported values are means over the five runs, and $\pm$ denotes the standard deviation across these runs, normalized by $n$. Accordingly, the $\text{MaxAE}$ reported in the tables is the mean of the five per-run maxima rather than the largest error observed across all runs.

\section{Physics-Constrained Loss Training Strategies}\label{app:pf_loss_training_ablation}

To evaluate the optimal strategy for integrating the AC power balance equations into the overall training objective, we conduct an ablation study comparing several physics-informed training regimes. A fundamental challenge in this formulation is the scale disparity between the supervised metric convergence ($\approx 10^{-5}$) and the purely unsupervised mean nodal power mismatch ($\approx 10^{2}$). Naively summing these objectives causes the physics gradients to dominate, preventing the optimizer from utilizing the supervised data. To address this, we compare an unsupervised baseline against multiple statically weighted formulations ($\lambda_{phys} \in \{10^{-6}, 10^{-7}, 10^{-8}, 10^{-9}\}$).

Because the stiffness of the AC power flow equations makes static weighting challenging to tune, we also evaluate a fine-tuning strategy. This approach trains the model to supervised convergence first, followed by structural refinement for physical feasibility. We evaluate this fine-tuning using both a purely unsupervised physics loss and our weighted physics losses. All variants train until convergence or a maximum of 4000 epochs.
For the fine-tuning variants, up to 10\% of training (400 epochs) is allocated for refinement. The performance trade-offs are summarized in Table \ref{tab:pfloss_training_strategy_ablation}.
The evaluation was conducted on the in-distribution validation split, which contains samples of all ten feeders, and the best configuration was then frozen and applied unchanged to all eleven test scenarios.

\begin{table}[t]
\centering
\caption{Training strategies for physics-constrained loss}
\label{tab:pfloss_training_strategy_ablation}
\setlength{\tabcolsep}{4pt}
\begin{tabular}{lcccc}
\toprule
\textbf{Model} & $\bm{\lambda_{phys}}$ & \multicolumn{3}{c}{\textbf{ID (Validation Set)}} \\
 & & \textbf{VM$_{\text{(p.u.)}}$} & \textbf{VA$_{\text{(deg)}}$} & \textbf{Train$_{\text{(s)}}$} \\
\midrule
Unsupervised & / & 1.0389 & 154.7869 & 20677.5 \\
Weighted & ($10^{-6}$) & 0.0144 & 0.2832 & 18857.5 \\
Weighted & ($10^{-7}$) & 0.0077 & 0.2746 & 18899.5 \\
Weighted & ($10^{-8}$) & 0.0049 & 0.1036 & 18862.6 \\
Weighted & ($10^{-9}$) & \textbf{0.0043} & 0.1510 & 18949.9 \\
Fine-tuning & / & 0.0075 & 0.2032 & \textbf{3552.5} \\
Fine-tuning & $10^{-7}$ & 0.0058 & 0.1210 & 3573.2 \\
Fine-tuning & $10^{-8}$ & \textbf{0.0043} & \textbf{0.0908} & 3566.8 \\
Fine-tuning & $10^{-9}$ & 0.0046 & 0.1289 & 3561.3 \\
\bottomrule
\end{tabular}
\end{table}

As observed, purely unsupervised training fails to identify a stable optimization basin, yielding severely high error across both state variable predictions. This indicates that relying solely on the nonlinear AC power balance equations without data-driven anchoring is insufficient for reliable power flow estimation. Furthermore, incorporating the physics loss from the beginning of the training cycle imposes a significant computational penalty. Models trained from scratch require approximately $5.5$ hours to converge, whereas the two-stage fine-tuning strategies complete in $\approx 58$ minutes. This reduction in computational overhead demonstrates the efficiency of initially navigating the loss landscape using the smoother, supervised objective.

The two-stage fine-tuning approach achieves the best balance of accuracy and efficiency. However, fine-tuning with a strictly unsupervised physics loss degraded performance compared to the weighted variants. This suggests that during the refinement phase, the optimizer still requires a supervised reference to maintain the learned state variable distributions. By employing a lightly weighted physics loss during fine-tuning, the model successfully maintains data fidelity while adjusting its weights toward physical feasibility. The fine-tuning formulation utilizing $\lambda_{phys} = 10^{-8}$ yields the lowest errors for voltage magnitude ($0.0043$ p.u.) and voltage angle ($0.0908^{\circ}$), while also being among the less computationally demanding variants. Consequently, we adopt this two-stage training paradigm for the \texttt{PFLoss} model in all subsequent experiments to effectively navigate the stiff loss landscape imposed by distribution grid constraints.

\section{Complex-Valued Model Implementation} \label{app:cplx}

Complex layers use the \texttt{complexPyTorch} library \citep{complexPyTorch2021}; message passing, batch normalization, and the readout are custom implementations built on PyTorch Geometric's \emph{MessagePassing} interface.

\textbf{Encoding.} Real-valued inputs are mapped to $\mathbb{C}$ once, before the network: $s_i = p_i + jq_i$, $z_{ij} = r_{ij} + jx_{ij}$, $V_{\text{slack}} = |V_{\text{slack}}|e^{j\theta_{\text{slack}}}$, and $z_{\text{slack}} = r_{\text{slack}} + jx_{\text{slack}}$. The hop distance to the slack bus is real and is embedded with zero imaginary part.

\textbf{Architecture.} Two complex fully connected preprocessing layers ($4 \rightarrow 256 \rightarrow 128$), seven complex message-passing layers at 128 complex channels, two complex postprocessing layers ($128 \rightarrow 256$), and a complex readout to a single complex output. Each message-passing layer and each pre/postprocessing layer is followed by complex batch normalization and $\mathbb{C}$ReLU.

\textbf{Optimization.} Complex parameters are stored as separate real and imaginary tensors, so gradients follow PyTorch's real-valued autograd on those components, equivalent to Wirtinger calculus for the real-valued objective used here. Training uses the same optimizer, learning-rate schedule, batch size, epoch budget, and seeds as the real-valued variants (Section~\ref{sec:evaluation_procedure}).

\section{Physical Consistency Metrics}\label{app:consistency}

\subsection{Nodal Apparent Power Mismatch}

Given predicted voltages $\hat{V}$, the residuals $L_{P_i}$ and $L_{Q_i}$ of Equations~\ref{eq:active_power}--\ref{eq:reactive_power} give the apparent power mismatch at bus $i$ as $S_i = \sqrt{L_{P_i}^{2} + L_{Q_i}^{2}}$, from which $S_{\text{nodal}}$ is the mean over all PQ buses in the graph.

This quantity is fairly common in the transmission literature \citep{rivera2025pf} but is highly sensitive to spatial voltage errors in distribution networks. Distribution feeders consist of short cables with small series impedance, so that, through $I = \Delta V / Z$, a small error in the voltage difference between adjacent buses implies a large error in the branch current and hence in the nodal power balance. Large values of $S_{\text{nodal}}$ therefore reflect physically meaningful violations of the network equations, but they may be driven by errors in the voltage differences between adjacent buses rather than by errors in the voltages themselves, and can be large even when nodal voltage errors are within measurement tolerance. Section~\ref{res:physical_consistency} illustrates this directly: a simulated meter operating within its rated tolerance produces mismatches several times larger than any learned model.

\subsection{Substation Loading}

As a complementary measure, we isolate the mismatch at the slack bus, $S_{\text{slack}} = \sqrt{(\Delta P_{\text{slack}})^{2} + (\Delta Q_{\text{slack}})^{2}}$. In distribution networks the slack bus is the primary substation transformer. Its leakage reactance is substantially larger than the impedance of downstream cables, which reduces the sensitivity described above. It is also the operationally meaningful boundary: the net power drawn from or exported to the upstream transmission grid is the quantity operators monitor for capacity planning and thermal limits.

\subsection{Normalization}

Both metrics are normalized by feeder throughput to remove the scale of feeder loading, which spans $0.17$--$52.55$~MW (Table~\ref{tab:engage_stats}). As $S_{\text{nodal}}$ is a mean over buses, its normalized value is not invariant to feeder size, so that values pooled over heterogeneous feeders depend on their composition. All variants are evaluated on identical test networks and thus under the same composition, but the normalized values themselves should not be compared across feeders of different size. Summing signed apparent power injections across the grid is unsuitable as a scaling factor, since generation approximately balances demand and the sum approaches zero, artificially inflating the normalized error. Summing load buses alone is also insufficient in active distribution networks, where high DER penetration produces intervals of negative net demand. We therefore scale by the maximum gross throughput,
\[
S_{\text{gross}} = \max\!\left(\left|\textstyle\sum_{i \in \mathcal{C}} S_i\right|,\ \left|\textstyle\sum_{i \in \mathcal{P}} S_i\right|\right),
\]
where $S_i$ is the net complex power injection of non-slack bus $i$, and $\mathcal{C}$ and $\mathcal{P}$ are the sets of non-slack buses with net active power consumption and production, respectively, giving $S_{\text{slack}} / S_{\text{gross}}$ and the analogous normalization for $S_{\text{nodal}}$. This avoids deflation of the scaling factor, limits the influence of line losses, and reflects the operational scale of both importing and exporting states.

\begin{table*}[t!]
\centering
\caption{Comparison of the evaluated variants against two published GNN power-flow models.}
\label{tab:gnn_ablation_study}
\setlength{\tabcolsep}{4pt}
\begin{tabular}{lcccc}
\toprule
\textbf{Model} & \multicolumn{2}{c}{\textbf{ID (All)}} & \multicolumn{2}{c}{\textbf{OOD}} \\
 & \textbf{VM$_{\text{(p.u.)}}$} & \textbf{VA$_{\text{(deg)}}$} & \textbf{VM$_{\text{(p.u.)}}$} & \textbf{VA$_{\text{(deg)}}$} \\
\midrule
Base & 0.0045 {\scriptsize $\pm$ 0.0007} & 0.1254 {\scriptsize $\pm$ 0.0357} & 0.0154 {\scriptsize $\pm$ 0.0011} & 1.0060 {\scriptsize $\pm$ 0.0467} \\
Cplx & \textbf{0.0026 {\scriptsize $\pm$ 0.0002}} & \textbf{0.0837 {\scriptsize $\pm$ 0.0114}} & 0.0158 {\scriptsize $\pm$ 0.0012} & \textbf{0.8023 {\scriptsize $\pm$ 0.0346}} \\
PFLoss & 0.0044 {\scriptsize $\pm$ 0.0004} & 0.1062 {\scriptsize $\pm$ 0.0244} & 0.0155 {\scriptsize $\pm$ 0.0008} & 1.0077 {\scriptsize $\pm$ 0.1691} \\
Res & 0.0042 {\scriptsize $\pm$ 0.0004} & 0.0949 {\scriptsize $\pm$ 0.0036} & \textbf{0.0137 {\scriptsize $\pm$ 0.0013}} & 1.0617 {\scriptsize $\pm$ 0.1040} \\
D-GCN & 0.0081 {\scriptsize $\pm$ 0.0007} & 0.2843 {\scriptsize $\pm$ 0.0199} & 0.0147 {\scriptsize $\pm$ 0.0033} & 1.9235 {\scriptsize $\pm$ 0.0723} \\
PFNet & 0.0597 {\scriptsize $\pm$ 0.0398} & 9.0062 {\scriptsize $\pm$ 6.6802} & 0.0856 {\scriptsize $\pm$ 0.0118} & 12.6015 {\scriptsize $\pm$ 2.1238} \\
\bottomrule
\end{tabular}
\end{table*}

\subsection{Simulated Smart Meter}\label{app:sensor}

The sensor reported in Figures~\ref{fig:performance_summary_gnn_models} and \ref{fig:physical_consistency_comparison} perturbs the ground-truth state with zero-mean noise scaled to the manufacturer's stated tolerances of $1\%$ for voltage magnitude and $1^{\circ}$ for voltage angle \citep{IMSys_ref}. It is included as a reference point for the accuracy of instrumentation currently available in distribution networks, not as a competing method. Because it perturbs the true state rather than predicting it, its error reflects only the sensitivity of each metric to small state perturbations.

We treat the stated tolerance as an expanded uncertainty with coverage factor $k=2$, corresponding to a $95\%$ interval, following the convention described in \citet{joint2008evaluation}. This is a modelling assumption rather than a property of the specification. Were the tolerance instead interpreted as a maximum permissible error, a rectangular distribution over the stated bounds would be the appropriate treatment \citep[\S4.3.7]{joint2008evaluation}, yielding a standard uncertainty approximately $15\%$ larger and correspondingly larger simulated measurement errors. Our choice is therefore the conservative one wherever this reference is used. It produces a more accurate simulated meter, against which our models are compared, and a smaller apparent power mismatch, understating rather than overstating the sensitivity we report.

Two further caveats apply. A meter's phase-angle tolerance characterizes a locally measured phase relationship rather than a bus voltage angle referenced to the slack, so we adopt $1^{\circ}$ as an illustrative reference magnitude only. The simulated meter is also idealized relative to deployed advanced metering infrastructure, which is additionally subject to latency and to missing or corrupted values \citep{cheng2023survey}, making it an optimistic reference.

Finally, we verify that our implementation returns negligible mismatch when evaluated on the ground-truth voltages themselves, confirming that the reported values arise from the predicted states rather than from the implementation.

\begin{table}[!t]
\centering
\caption{Structural dissimilarity of each held-out feeder from its corresponding training pool, measured by Maximum Mean Discrepancy (MMD).
Higher values indicate a test topology less well represented by the grids the model was trained on. Bold indicates the largest value in each column.}
\label{tab:ood_mmds}
\small
\setlength{\tabcolsep}{5pt}
\begin{tabular}{@{}lccc@{}}
\toprule
\textbf{Held-out grid} & \textbf{Buses} & \textbf{MMD}$_{\mathrm{degree}}$ & \textbf{MMD}$_{\mathrm{Laplacian}}$ \\
\midrule
LV-rural1   & 15  & \textbf{1.4209} & \textbf{1.3344} \\
LV-rural2   & 97  & 0.8309 & 0.7993 \\
LV-rural3   & 129 & 0.6801 & 0.6718 \\
LV-semiurb4 & 44  & 0.7555 & 1.0931 \\
LV-semiurb5 & 111 & 0.7364 & 0.7450 \\
LV-urban6   & 59  & 1.0571 & 0.9677 \\
MV-comm     & 106 & 0.4687 & 0.6428 \\
MV-rural    & 97  & 0.4957 & 0.7097 \\
MV-semiurb  & 120 & 0.5113 & 0.6836 \\
MV-urban    & 136 & 0.9328 & 0.7418 \\
\bottomrule
\end{tabular}
\end{table}

\section{Comparison with Learned Baselines} \label{app:learned_baselines}

Our study compares inductive biases under a fixed backbone. To verify that conclusions drawn in this setting are not artifacts of an unrepresentative neural architecture, we benchmark our variants against two learned baselines from the recent literature.

We denote \textbf{D-GCN} as the degree-augmented graph convolutional network introduced in the ENGAGE generalization framework \citep{okoyomon2025framework}, where it was the strongest performing architecture across the same ten distribution feeders used here. \textbf{PFNet} is PowerFlowNet \citep{lin_powerflownet_2024}, a message-passing GNN for power flow approximation, built on K-hop TAGConv layers. We use the open-source implementation provided with the PF$\Delta$ benchmark \citep{rivera2025pf}. Both models were adapted to the ENGAGE data structure and received a hyperparameter tuning similar to our own variants, covering number of layers, learning rate, epoch budget, and loss function. All models were trained across the same five seeds and evaluated under the identical in-distribution and leave-one-feeder-out protocols.

Results are reported in Table~\ref{tab:gnn_ablation_study}. The Base architecture outperforms both comparators on three of four metrics, and the lower out-of-distribution voltage magnitude error of D-GCN ($0.0147 \pm 0.0033$ p.u.) is not distinguishable from that of our variants given the seed variance. This supports the use of the Base model as a representative control for the comparisons in Section~\ref{sec:results}.

We note that \textbf{PFNet} was developed and validated on a different data regime, and its performance here should not be read as characterizing the method. The large seed-to-seed variance we observe ($\pm 6.68^{\circ}$ for ID voltage angle) indicates training instability under our heterogeneous LV/MV setting rather than a converged comparison. We include it for completeness and draw no conclusions about its relative merit.

\begin{table*}[b]
\centering
\caption{Experimental results of physics-informed GNN variants and analytical baselines across all eleven test scenarios ($n=5$). MaxAE is the mean of the five per-run maxima, and $\pm$ denotes the standard deviation across runs.}
\label{tab:model_performance_raw}
\begin{tabular}{llcccc}
\toprule
\textbf{Testing Grid} & \textbf{Model} & \multicolumn{2}{c}{\textbf{VM$_{\text{(p.u.)}}$}} & \multicolumn{2}{c}{\textbf{VA$_{\text{(deg)}}$}} \\
 &  & \textit{MAE$^{\text{Median}}$} \tiny{[Q1, Q3]} & \textit{MaxAE} & \textit{MAE$^{\text{Median}}$} \tiny{[Q1, Q3]} & \textit{MaxAE} \\
\midrule
ID (All) & Base & 0.0045 {\tiny $\pm$ 0.0007} {\scriptsize [0.0032, 0.0064]} & 0.0409 {\tiny $\pm$ 0.0052} & 0.1254 {\tiny $\pm$ 0.0357} {\scriptsize [0.0859, 0.1819]} & 1.6814 {\tiny $\pm$ 0.4069} \\
ID (All) & Cplx & 0.0026 {\tiny $\pm$ 0.0002} {\scriptsize [0.0018, 0.0045]} & 0.0291 {\tiny $\pm$ 0.0051} & 0.0837 {\tiny $\pm$ 0.0114} {\scriptsize [0.0618, 0.1224]} & 1.7508 {\tiny $\pm$ 0.1702} \\
ID (All) & PFLoss & 0.0044 {\tiny $\pm$ 0.0004} {\scriptsize [0.0032, 0.0062]} & 0.0531 {\tiny $\pm$ 0.0323} & 0.1062 {\tiny $\pm$ 0.0244} {\scriptsize [0.0771, 0.1513]} & 2.1320 {\tiny $\pm$ 1.4375} \\
ID (All) & Res & 0.0042 {\tiny $\pm$ 0.0004} {\scriptsize [0.0031, 0.0057]} & 0.0384 {\tiny $\pm$ 0.0036} & 0.0949 {\tiny $\pm$ 0.0036} {\scriptsize [0.0722, 0.1375]} & 2.8113 {\tiny $\pm$ 2.5069} \\
ID (All) & DCPF & 0.0137 {\tiny $\pm$ 0.0004} {\scriptsize [0.0099, 0.0194]} & 0.0552 {\tiny $\pm$ 0.0013} & 0.2712 {\tiny $\pm$ 0.0087} {\scriptsize [0.1582, 0.3893]} & 1.8730 {\tiny $\pm$ 0.1552} \\
ID (All) & LDF & 0.0010 {\tiny $\pm$ 0.0000} {\scriptsize [0.0006, 0.0017]} & 0.0073 {\tiny $\pm$ 0.0000} & 0.0292 {\tiny $\pm$ 0.0004} {\scriptsize [0.0150, 0.0424]} & 0.3774 {\tiny $\pm$ 0.0193} \\
\midrule
LV-rural1 & Base & 0.0470 {\tiny $\pm$ 0.0271} {\scriptsize [0.0460, 0.0479]} & 0.1249 {\tiny $\pm$ 0.0761} & 7.2003 {\tiny $\pm$ 7.7054} {\scriptsize [7.1770, 7.2240]} & 21.0029 {\tiny $\pm$ 25.0517} \\
LV-rural1 & Cplx & 0.1242 {\tiny $\pm$ 0.0628} {\scriptsize [0.1235, 0.1249]} & 0.3742 {\tiny $\pm$ 0.2538} & 4.1821 {\tiny $\pm$ 1.0123} {\scriptsize [4.1512, 4.2139]} & 12.1068 {\tiny $\pm$ 3.3234} \\
LV-rural1 & PFLoss & 0.0549 {\tiny $\pm$ 0.0263} {\scriptsize [0.0544, 0.0555]} & 0.1459 {\tiny $\pm$ 0.0517} & 3.7216 {\tiny $\pm$ 1.6252} {\scriptsize [3.6991, 3.7439]} & 9.5063 {\tiny $\pm$ 3.3810} \\
LV-rural1 & Res & 0.0210 {\tiny $\pm$ 0.0089} {\scriptsize [0.0202, 0.0219]} & 0.0632 {\tiny $\pm$ 0.0211} & 4.2079 {\tiny $\pm$ 1.2103} {\scriptsize [4.1781, 4.2385]} & 14.5505 {\tiny $\pm$ 9.3872} \\
LV-rural1 & DCPF & 0.0113 {\tiny $\pm$ 0.0000} {\scriptsize [0.0099, 0.0126]} & 0.0314 {\tiny $\pm$ 0.0000} & 0.0911 {\tiny $\pm$ 0.0000} {\scriptsize [0.0569, 0.1306]} & 0.4578 {\tiny $\pm$ 0.0000} \\
LV-rural1 & LDF & 0.0010 {\tiny $\pm$ 0.0000} {\scriptsize [0.0010, 0.0010]} & 0.0011 {\tiny $\pm$ 0.0000} & 0.0324 {\tiny $\pm$ 0.0000} {\scriptsize [0.0297, 0.0351]} & 0.0502 {\tiny $\pm$ 0.0000} \\
\midrule
LV-rural2 & Base & 0.0208 {\tiny $\pm$ 0.0035} {\scriptsize [0.0179, 0.0242]} & 0.0730 {\tiny $\pm$ 0.0148} & 1.4258 {\tiny $\pm$ 0.3231} {\scriptsize [1.3299, 1.5075]} & 8.2236 {\tiny $\pm$ 4.4573} \\
LV-rural2 & Cplx & 0.0201 {\tiny $\pm$ 0.0030} {\scriptsize [0.0186, 0.0219]} & 0.1054 {\tiny $\pm$ 0.0492} & 1.0905 {\tiny $\pm$ 0.2604} {\scriptsize [0.9959, 1.1690]} & 4.8516 {\tiny $\pm$ 1.2149} \\
LV-rural2 & PFLoss & 0.0169 {\tiny $\pm$ 0.0077} {\scriptsize [0.0142, 0.0204]} & 0.0680 {\tiny $\pm$ 0.0270} & 1.5515 {\tiny $\pm$ 0.8877} {\scriptsize [1.4575, 1.6306]} & 6.2372 {\tiny $\pm$ 6.3128} \\
LV-rural2 & Res & 0.0196 {\tiny $\pm$ 0.0076} {\scriptsize [0.0177, 0.0219]} & 0.0903 {\tiny $\pm$ 0.0269} & 1.8192 {\tiny $\pm$ 0.1292} {\scriptsize [1.7347, 1.8935]} & 9.8442 {\tiny $\pm$ 2.7719} \\
LV-rural2 & DCPF & 0.0273 {\tiny $\pm$ 0.0000} {\scriptsize [0.0215, 0.0322]} & 0.0474 {\tiny $\pm$ 0.0000} & 0.4787 {\tiny $\pm$ 0.0000} {\scriptsize [0.3779, 0.5989]} & 1.4663 {\tiny $\pm$ 0.0000} \\
LV-rural2 & LDF & 0.0004 {\tiny $\pm$ 0.0000} {\scriptsize [0.0002, 0.0006]} & 0.0011 {\tiny $\pm$ 0.0000} & 0.0101 {\tiny $\pm$ 0.0000} {\scriptsize [0.0085, 0.0123]} & 0.0267 {\tiny $\pm$ 0.0000} \\
\midrule
\multicolumn{6}{c}{\itshape\footnotesize $\cdots$ continued on next page $\cdots$} \\
\bottomrule
\end{tabular}
\end{table*}

\begin{table*}[t]
\ContinuedFloat
\centering
\caption{Experimental results of physics-informed GNN variants and analytical baselines across all eleven test scenarios ($n=5$). MaxAE is the mean of the five per-run maxima, and $\pm$ denotes the standard deviation across runs.}
\begin{tabular}{llcccc}
\toprule
\multicolumn{6}{c}{\itshape\footnotesize $\cdots$ continued from previous page $\cdots$} \\
\midrule
\textbf{Testing Grid} & \textbf{Model} & \multicolumn{2}{c}{\textbf{VM$_{\text{(p.u.)}}$}} & \multicolumn{2}{c}{\textbf{VA$_{\text{(deg)}}$}} \\
 &  & \textit{MAE$^{\text{Median}}$} \tiny{[Q1, Q3]} & \textit{MaxAE} & \textit{MAE$^{\text{Median}}$} \tiny{[Q1, Q3]} & \textit{MaxAE} \\
\midrule
LV-rural3 & Base & 0.0169 {\tiny $\pm$ 0.0025} {\scriptsize [0.0158, 0.0188]} & 0.1548 {\tiny $\pm$ 0.0519} & 0.9180 {\tiny $\pm$ 0.1484} {\scriptsize [0.7885, 1.0638]} & 10.2734 {\tiny $\pm$ 2.8338} \\
LV-rural3 & Cplx & 0.0268 {\tiny $\pm$ 0.0097} {\scriptsize [0.0255, 0.0296]} & 0.6428 {\tiny $\pm$ 0.2646} & 0.9989 {\tiny $\pm$ 0.1832} {\scriptsize [0.8670, 1.1444]} & 10.1793 {\tiny $\pm$ 1.8965} \\
LV-rural3 & PFLoss & 0.0171 {\tiny $\pm$ 0.0041} {\scriptsize [0.0160, 0.0188]} & 0.1665 {\tiny $\pm$ 0.0746} & 0.9548 {\tiny $\pm$ 0.1707} {\scriptsize [0.8285, 1.1026]} & 14.9659 {\tiny $\pm$ 7.8663} \\
LV-rural3 & Res & 0.0154 {\tiny $\pm$ 0.0014} {\scriptsize [0.0142, 0.0171]} & 0.0935 {\tiny $\pm$ 0.0354} & 0.9302 {\tiny $\pm$ 0.0972} {\scriptsize [0.7904, 1.0811]} & 10.2739 {\tiny $\pm$ 9.9438} \\
LV-rural3 & DCPF & 0.0254 {\tiny $\pm$ 0.0000} {\scriptsize [0.0184, 0.0325]} & 0.0576 {\tiny $\pm$ 0.0000} & 0.5828 {\tiny $\pm$ 0.0000} {\scriptsize [0.4387, 0.7511]} & 2.0496 {\tiny $\pm$ 0.0000} \\
LV-rural3 & LDF & 0.0011 {\tiny $\pm$ 0.0000} {\scriptsize [0.0007, 0.0016]} & 0.0031 {\tiny $\pm$ 0.0000} & 0.0165 {\tiny $\pm$ 0.0000} {\scriptsize [0.0109, 0.0234]} & 0.0458 {\tiny $\pm$ 0.0000} \\
\midrule
LV-semiurb4 & Base & 0.0159 {\tiny $\pm$ 0.0016} {\scriptsize [0.0147, 0.0174]} & 0.0444 {\tiny $\pm$ 0.0024} & 0.7944 {\tiny $\pm$ 0.0682} {\scriptsize [0.7388, 0.8448]} & 1.7103 {\tiny $\pm$ 0.4303} \\
LV-semiurb4 & Cplx & 0.0151 {\tiny $\pm$ 0.0013} {\scriptsize [0.0140, 0.0162]} & 0.0467 {\tiny $\pm$ 0.0066} & 0.6667 {\tiny $\pm$ 0.0334} {\scriptsize [0.6122, 0.7167]} & 1.5324 {\tiny $\pm$ 0.1221} \\
LV-semiurb4 & PFLoss & 0.0160 {\tiny $\pm$ 0.0014} {\scriptsize [0.0148, 0.0175]} & 0.0444 {\tiny $\pm$ 0.0053} & 0.7843 {\tiny $\pm$ 0.0693} {\scriptsize [0.7288, 0.8348]} & 1.4457 {\tiny $\pm$ 0.1395} \\
LV-semiurb4 & Res & 0.0147 {\tiny $\pm$ 0.0034} {\scriptsize [0.0135, 0.0162]} & 0.0441 {\tiny $\pm$ 0.0091} & 0.7701 {\tiny $\pm$ 0.1240} {\scriptsize [0.7145, 0.8205]} & 1.4734 {\tiny $\pm$ 0.1745} \\
LV-semiurb4 & DCPF & 0.0189 {\tiny $\pm$ 0.0000} {\scriptsize [0.0178, 0.0202]} & 0.0460 {\tiny $\pm$ 0.0000} & 0.2614 {\tiny $\pm$ 0.0000} {\scriptsize [0.2160, 0.3014]} & 0.7923 {\tiny $\pm$ 0.0000} \\
LV-semiurb4 & LDF & 0.0007 {\tiny $\pm$ 0.0000} {\scriptsize [0.0006, 0.0009]} & 0.0015 {\tiny $\pm$ 0.0000} & 0.0376 {\tiny $\pm$ 0.0000} {\scriptsize [0.0304, 0.0459]} & 0.0550 {\tiny $\pm$ 0.0000} \\
\midrule
LV-semiurb5 & Base & 0.0227 {\tiny $\pm$ 0.0110} {\scriptsize [0.0209, 0.0255]} & 0.1061 {\tiny $\pm$ 0.0507} & 0.5889 {\tiny $\pm$ 0.0976} {\scriptsize [0.5412, 0.6510]} & 2.9217 {\tiny $\pm$ 0.6040} \\
LV-semiurb5 & Cplx & 0.0147 {\tiny $\pm$ 0.0011} {\scriptsize [0.0129, 0.0174]} & 0.0660 {\tiny $\pm$ 0.0140} & 0.6022 {\tiny $\pm$ 0.0940} {\scriptsize [0.5455, 0.6713]} & 2.6963 {\tiny $\pm$ 0.5920} \\
LV-semiurb5 & PFLoss & 0.0163 {\tiny $\pm$ 0.0029} {\scriptsize [0.0147, 0.0187]} & 0.0791 {\tiny $\pm$ 0.0292} & 0.6896 {\tiny $\pm$ 0.1537} {\scriptsize [0.6536, 0.7422]} & 3.3718 {\tiny $\pm$ 0.8127} \\
LV-semiurb5 & Res & 0.0145 {\tiny $\pm$ 0.0029} {\scriptsize [0.0130, 0.0173]} & 0.0630 {\tiny $\pm$ 0.0067} & 0.6083 {\tiny $\pm$ 0.1467} {\scriptsize [0.5630, 0.6674]} & 4.0524 {\tiny $\pm$ 2.9397} \\
LV-semiurb5 & DCPF & 0.0120 {\tiny $\pm$ 0.0000} {\scriptsize [0.0106, 0.0157]} & 0.0452 {\tiny $\pm$ 0.0000} & 0.4358 {\tiny $\pm$ 0.0000} {\scriptsize [0.3291, 0.5282]} & 1.2760 {\tiny $\pm$ 0.0000} \\
LV-semiurb5 & LDF & 0.0009 {\tiny $\pm$ 0.0000} {\scriptsize [0.0006, 0.0013]} & 0.0033 {\tiny $\pm$ 0.0000} & 0.0196 {\tiny $\pm$ 0.0000} {\scriptsize [0.0135, 0.0256]} & 0.0456 {\tiny $\pm$ 0.0000} \\
\midrule
LV-urban6 & Base & 0.0097 {\tiny $\pm$ 0.0013} {\scriptsize [0.0090, 0.0105]} & 0.0378 {\tiny $\pm$ 0.0085} & 0.6094 {\tiny $\pm$ 0.0347} {\scriptsize [0.5858, 0.6555]} & 1.5512 {\tiny $\pm$ 0.2130} \\
LV-urban6 & Cplx & 0.0093 {\tiny $\pm$ 0.0009} {\scriptsize [0.0087, 0.0101]} & 0.0395 {\tiny $\pm$ 0.0033} & 0.6338 {\tiny $\pm$ 0.0613} {\scriptsize [0.6163, 0.6589]} & 1.7641 {\tiny $\pm$ 0.3483} \\
LV-urban6 & PFLoss & 0.0088 {\tiny $\pm$ 0.0007} {\scriptsize [0.0082, 0.0096]} & 0.0343 {\tiny $\pm$ 0.0025} & 0.6145 {\tiny $\pm$ 0.0030} {\scriptsize [0.5951, 0.6457]} & 1.8103 {\tiny $\pm$ 0.7306} \\
LV-urban6 & Res & 0.0082 {\tiny $\pm$ 0.0006} {\scriptsize [0.0075, 0.0092]} & 0.0359 {\tiny $\pm$ 0.0034} & 0.6147 {\tiny $\pm$ 0.0779} {\scriptsize [0.5943, 0.6500]} & 1.3273 {\tiny $\pm$ 0.1802} \\
LV-urban6 & DCPF & 0.0111 {\tiny $\pm$ 0.0000} {\scriptsize [0.0098, 0.0135]} & 0.0353 {\tiny $\pm$ 0.0000} & 0.2634 {\tiny $\pm$ 0.0000} {\scriptsize [0.2176, 0.3097]} & 0.8363 {\tiny $\pm$ 0.0000} \\
LV-urban6 & LDF & 0.0006 {\tiny $\pm$ 0.0000} {\scriptsize [0.0004, 0.0008]} & 0.0017 {\tiny $\pm$ 0.0000} & 0.0220 {\tiny $\pm$ 0.0000} {\scriptsize [0.0162, 0.0279]} & 0.0373 {\tiny $\pm$ 0.0000} \\
\midrule
MV-comm & Base & 0.0084 {\tiny $\pm$ 0.0009} {\scriptsize [0.0072, 0.0104]} & 0.0726 {\tiny $\pm$ 0.0209} & 1.1192 {\tiny $\pm$ 0.2413} {\scriptsize [1.0858, 1.1581]} & 6.0982 {\tiny $\pm$ 5.6492} \\
MV-comm & Cplx & 0.0070 {\tiny $\pm$ 0.0009} {\scriptsize [0.0059, 0.0090]} & 0.1416 {\tiny $\pm$ 0.0770} & 0.7740 {\tiny $\pm$ 0.1252} {\scriptsize [0.7205, 0.8555]} & 7.2760 {\tiny $\pm$ 1.4939} \\
MV-comm & PFLoss & 0.0080 {\tiny $\pm$ 0.0014} {\scriptsize [0.0068, 0.0099]} & 0.0557 {\tiny $\pm$ 0.0162} & 1.1658 {\tiny $\pm$ 0.2763} {\scriptsize [1.1391, 1.1929]} & 3.5310 {\tiny $\pm$ 1.4311} \\
MV-comm & Res & 0.0083 {\tiny $\pm$ 0.0010} {\scriptsize [0.0076, 0.0094]} & 0.0760 {\tiny $\pm$ 0.0133} & 1.2578 {\tiny $\pm$ 0.1937} {\scriptsize [1.2325, 1.2953]} & 4.4966 {\tiny $\pm$ 2.6414} \\
MV-comm & DCPF & 0.0063 {\tiny $\pm$ 0.0000} {\scriptsize [0.0041, 0.0111]} & 0.0441 {\tiny $\pm$ 0.0000} & 0.2098 {\tiny $\pm$ 0.0000} {\scriptsize [0.1693, 0.2582]} & 1.3392 {\tiny $\pm$ 0.0000} \\
MV-comm & LDF & 0.0028 {\tiny $\pm$ 0.0000} {\scriptsize [0.0025, 0.0030]} & 0.0085 {\tiny $\pm$ 0.0000} & 0.0397 {\tiny $\pm$ 0.0000} {\scriptsize [0.0343, 0.0475]} & 0.2433 {\tiny $\pm$ 0.0000} \\
\midrule
MV-rural & Base & 0.0157 {\tiny $\pm$ 0.0026} {\scriptsize [0.0147, 0.0168]} & 0.1338 {\tiny $\pm$ 0.0544} & 1.5205 {\tiny $\pm$ 0.6237} {\scriptsize [1.4818, 1.5628]} & 13.6617 {\tiny $\pm$ 4.9345} \\
MV-rural & Cplx & 0.0145 {\tiny $\pm$ 0.0017} {\scriptsize [0.0139, 0.0153]} & 0.0780 {\tiny $\pm$ 0.0162} & 0.7033 {\tiny $\pm$ 0.0952} {\scriptsize [0.6720, 0.7471]} & 3.0812 {\tiny $\pm$ 0.6597} \\
MV-rural & PFLoss & 0.0184 {\tiny $\pm$ 0.0059} {\scriptsize [0.0159, 0.0209]} & 0.1539 {\tiny $\pm$ 0.0731} & 1.6464 {\tiny $\pm$ 0.5021} {\scriptsize [1.6237, 1.6754]} & 12.9819 {\tiny $\pm$ 5.5285} \\
MV-rural & Res & 0.0162 {\tiny $\pm$ 0.0025} {\scriptsize [0.0151, 0.0175]} & 0.1120 {\tiny $\pm$ 0.0345} & 1.2155 {\tiny $\pm$ 0.3034} {\scriptsize [1.1704, 1.2662]} & 12.7663 {\tiny $\pm$ 8.5436} \\
MV-rural & DCPF & 0.0134 {\tiny $\pm$ 0.0000} {\scriptsize [0.0098, 0.0174]} & 0.0577 {\tiny $\pm$ 0.0000} & 0.2518 {\tiny $\pm$ 0.0000} {\scriptsize [0.1824, 0.3649]} & 1.8193 {\tiny $\pm$ 0.0000} \\
MV-rural & LDF & 0.0051 {\tiny $\pm$ 0.0000} {\scriptsize [0.0049, 0.0052]} & 0.0074 {\tiny $\pm$ 0.0000} & 0.1293 {\tiny $\pm$ 0.0000} {\scriptsize [0.1274, 0.1325]} & 0.3685 {\tiny $\pm$ 0.0000} \\
\midrule
MV-semiurb & Base & 0.0112 {\tiny $\pm$ 0.0046} {\scriptsize [0.0100, 0.0125]} & 0.0984 {\tiny $\pm$ 0.0409} & 1.7888 {\tiny $\pm$ 0.0825} {\scriptsize [1.7472, 1.8297]} & 10.2876 {\tiny $\pm$ 6.9884} \\
MV-semiurb & Cplx & 0.0195 {\tiny $\pm$ 0.0097} {\scriptsize [0.0184, 0.0207]} & 0.2531 {\tiny $\pm$ 0.0953} & 1.8275 {\tiny $\pm$ 0.1843} {\scriptsize [1.7826, 1.8828]} & 12.9613 {\tiny $\pm$ 4.3020} \\
MV-semiurb & PFLoss & 0.0117 {\tiny $\pm$ 0.0025} {\scriptsize [0.0105, 0.0134]} & 0.1729 {\tiny $\pm$ 0.0990} & 1.6782 {\tiny $\pm$ 0.2495} {\scriptsize [1.6344, 1.7204]} & 10.3402 {\tiny $\pm$ 7.3244} \\
MV-semiurb & Res & 0.0116 {\tiny $\pm$ 0.0027} {\scriptsize [0.0106, 0.0127]} & 0.1079 {\tiny $\pm$ 0.0307} & 1.5964 {\tiny $\pm$ 0.2052} {\scriptsize [1.5617, 1.6336]} & 10.5327 {\tiny $\pm$ 4.0286} \\
MV-semiurb & DCPF & 0.0090 {\tiny $\pm$ 0.0000} {\scriptsize [0.0062, 0.0134]} & 0.0487 {\tiny $\pm$ 0.0000} & 0.2301 {\tiny $\pm$ 0.0000} {\scriptsize [0.1708, 0.3112]} & 2.3028 {\tiny $\pm$ 0.0000} \\
MV-semiurb & LDF & 0.0014 {\tiny $\pm$ 0.0000} {\scriptsize [0.0010, 0.0017]} & 0.0022 {\tiny $\pm$ 0.0000} & 0.0146 {\tiny $\pm$ 0.0000} {\scriptsize [0.0141, 0.0159]} & 0.0514 {\tiny $\pm$ 0.0000} \\
\midrule
MV-urban & Base & 0.0203 {\tiny $\pm$ 0.0041} {\scriptsize [0.0186, 0.0220]} & 0.1086 {\tiny $\pm$ 0.0275} & 0.9821 {\tiny $\pm$ 0.1394} {\scriptsize [0.9629, 1.0054]} & 8.4945 {\tiny $\pm$ 3.3797} \\
MV-urban & Cplx & 0.0174 {\tiny $\pm$ 0.0005} {\scriptsize [0.0164, 0.0184]} & 0.1269 {\tiny $\pm$ 0.0235} & 0.8090 {\tiny $\pm$ 0.0424} {\scriptsize [0.7911, 0.8294]} & 6.4527 {\tiny $\pm$ 3.6643} \\
MV-urban & PFLoss & 0.0179 {\tiny $\pm$ 0.0018} {\scriptsize [0.0157, 0.0205]} & 0.1002 {\tiny $\pm$ 0.0201} & 0.9134 {\tiny $\pm$ 0.1600} {\scriptsize [0.8944, 0.9334]} & 10.3538 {\tiny $\pm$ 3.5551} \\
MV-urban & Res & 0.0157 {\tiny $\pm$ 0.0038} {\scriptsize [0.0141, 0.0176]} & 0.1052 {\tiny $\pm$ 0.0252} & 0.9079 {\tiny $\pm$ 0.0942} {\scriptsize [0.8912, 0.9252]} & 10.9173 {\tiny $\pm$ 2.8213} \\
MV-urban & DCPF & 0.0125 {\tiny $\pm$ 0.0000} {\scriptsize [0.0074, 0.0182]} & 0.0335 {\tiny $\pm$ 0.0000} & 0.1069 {\tiny $\pm$ 0.0000} {\scriptsize [0.0814, 0.1351]} & 0.7344 {\tiny $\pm$ 0.0000} \\
MV-urban & LDF & 0.0005 {\tiny $\pm$ 0.0000} {\scriptsize [0.0004, 0.0006]} & 0.0041 {\tiny $\pm$ 0.0000} & 0.0478 {\tiny $\pm$ 0.0000} {\scriptsize [0.0410, 0.0563]} & 0.5133 {\tiny $\pm$ 0.0000} \\
\bottomrule
\end{tabular}
\end{table*}

\section{Topology-Level Analysis of OOD Experiments} \label{app:topology-ood}

\citep{okoyomon2025framework} present a structural analysis of grid similarity in the ENGAGE dataset. They employ Maximum Mean Discrepancy (MMD) \cite{gretton2012kernel} as a principled and expressive means of comparing distributions of topology-level properties between graph datasets. Using this method, they find that the LV-rural1 subset of the dataset is the most different test case based on spectral properties and degree distribution, based on pairwise comparisons \citep[Figure 4]{okoyomon2025framework}. We use this same OOD analysis to quantify the difference between OOD test cases in Table~\ref{tab:ood_mmds}. For each leave-one-out scenario we compute the MMD between the graphs of the held-out feeder and the pooled graphs of the nine feeders used for training, over two topology-level feature distributions: the node degree distribution and the Laplacian spectrum. LV-rural1 is the most dissimilar held-out feeder under both measures, exceeding the next-highest grid by 34\% in degree divergence and 22\% in spectral divergence. This is not merely a consequence of its size: MV-urban, the largest feeder in the dataset, is more structurally dissimilar from its training pool than LV-rural3, which has nearly as many buses. The leave-one-out scenario for LV-rural1 is therefore the most demanding generalization task in our suite, which is consistent with it producing the worst median performance across all variants.

\section{Complete Benchmarking Results} \label{app:complete_benchmarking_results}

We present the full benchmarking results in Table \ref{tab:model_performance_raw}, which displays the raw performance metrics for all inductive bias models across all test scenarios.

%% --- Setup for Supplemental Materials ---
\clearpage
\newpage
\section*{Supplemental Materials}

% Reset the section, figure and table counters to 1
\setcounter{section}{0} 
\setcounter{figure}{0}
\setcounter{table}{0}

% Prefix sections, figures and tables with 'S'
\renewcommand{\thesection}{S\arabic{section}}
\renewcommand{\thefigure}{S\arabic{figure}}
\renewcommand{\thetable}{S\arabic{table}}

\section{ENGAGE Data Representation} \label{app:engage_data_representation}

Each graph in the ENGAGE dataset is represented as a PyTorch Geometric Data object. Since we are dealing with voltage prediction in distribution grids, there are only PQ buses and the slack bus. Additionally, we store the slack information globally and add the minimum hops to the slack bus for every bus as an additional feature. The model task is to use the bus information (\textit{data.x}), network connectivity (\textit{data.edge\_index}), line information (\textit{data.edge\_attr}), and global information (\textit{data.slack\_info}) to predict the output voltages at every bus (\textit{data.y}). To assist the physics-informed loss formulation, we additionally include the underlying PYPOWER case file information (\textit{data.ppci}) that was used to model the networks in pandapower. This allows us to correctly formulate our PyTorch power balance equation implementation, since different modeling frameworks make different network assumptions.

% \settablecounter{5}
\begin{table}[h!]
    \centering
    \caption{PyTorch Geometric Data Object Attributes}
    \label{tab:data_attributes}
    \begin{tabular}{|lcl|}
        \hline
        \textbf{Attribute} & \textbf{Shape} & \textbf{Description} \\
        \hline
        
        \texttt{x} & $(N, 3)$ & \begin{tabular}[t]{@{}l}Node features.\\ $N$ is the number of nodes/buses.\\ Features: \\ \quad - \texttt{p\_mw} \\ \quad - \texttt{q\_mvar} \\ \quad - \texttt{hops\_to\_slack} \end{tabular} \\
        \hline
        
        \texttt{edge\_index} & $(2, 2E)$ & \begin{tabular}[t]{@{}l}Graph connectivity in COO format.\\ $E$ is the number of lines.\\ Edges are directed.\end{tabular} \\
        \hline
        
        \texttt{edge\_attr} & $(2E, 2)$ & \begin{tabular}[t]{@{}l}Edge features.\\ $E$ is the number of lines.\\ Features: \\ \quad - \texttt{r\_pu} \\ \quad - \texttt{x\_pu} \end{tabular} \\
        \hline
        
        \texttt{slack\_info} & $(4)$ & \begin{tabular}[t]{@{}l}Static network-wide parameters.\\ Features: \\ \quad - \texttt{slack\_vm\_pu} \\ \quad - \texttt{slack\_va\_degree} \\ \quad - \texttt{slack\_r\_pu} \\ \quad - \texttt{slack\_x\_pu} \end{tabular} \\
        \hline
        
        \texttt{y} & $(N, 2)$ & \begin{tabular}[t]{@{}l}Target values for each node.\\ $N$ is the number of nodes/buses.\\ Labels: \\ \quad - \texttt{vm\_pu} \\ \quad - \texttt{va\_degree}\end{tabular} \\
        \hline
        
        \texttt{ppci} & N/A & \begin{tabular}[t]{@{}l}Internal pypower casefile used by \\ pandapower\tablefootnote{See \url{https://pandapower.readthedocs.io/en/v2.0.0/powerflow/ac.html}} for power flow. \\ (\texttt{net["\_ppc"]["internal"]})\end{tabular} \\
        \hline
    \end{tabular}
    \vspace{1em}
\end{table}

\section{Analytical Baselines} \label{app:analytical_baselines}

\subsection{DC Power Flow}

The DC power flow approximation linearizes the full AC method such that a non-iterative, convergent solution can be found. This simplification ignores reactive power and assumes small angle differences and a fixed voltage magnitude of 1.0 p.u. for all buses, and thus often leads to inaccurate predictions in situations with larger voltage drops.

\subsubsection{Formulation} \label{app:dcpf}

We utilize the implementation provided by the \textit{pandapower} library \citep{thurner2018pandapower}. The DCPF model is derived from the full AC power flow equations (Equations \ref{eq:active_power}-–\ref{eq:reactive_power}) by applying three simplifying assumptions:
\begin{enumerate}
    \item \textbf{Lossless Lines:} The line resistance is negligible ($r_{ij} \approx 0$), implying $G_{ij} \approx 0$.
    \item \textbf{Flat Voltage Profile:} Voltage magnitudes are fixed at $|v_i| \approx 1.0$ p.u. for all buses.
    \item \textbf{Small Voltage Angles:} The phase angle differences are small, such that $\sin(\theta_i - \theta_j) \approx \theta_i - \theta_j$ and $\cos(\theta_i - \theta_j) \approx 1$.
\end{enumerate}

Under these assumptions, the reactive power flows ($Q$) and voltage magnitude variations are decoupled and ignored. The system reduces to a set of linear equations relating active power injection $P_i$ to voltage angles $\theta$:
\begin{equation}
    p_{\text{net},i} = \sum_{j \in \mathcal{N}_i} B_{ij}(\theta_i - \theta_j) = \sum_{j \in \mathcal{N}_i} \frac{\theta_i - \theta_j}{x_{ij}}
\end{equation}

\subsection{LinDistFlow}

Linearized Distribution Flow (LinDistFlow) linearizes the non-linear AC power flow equations by neglecting second-order loss terms, providing a convex approximation suitable for voltage estimation in distribution grids. This formulation is derived from the standard DistFlow equations originally introduced by \citet{baran1989network} for radial distribution networks.

\subsubsection{Voltage Magnitude Estimation}
Consider a radial distribution network represented by a directed graph $\mathcal{G} = (\mathcal{N}, \mathcal{E})$, where $\mathcal{N}$ is the set of buses and $\mathcal{E}$ is the set of lines. For each line $(i, j) \in \mathcal{E}$ connecting bus $i$ to bus $j$, let $z_{ij} = r_{ij} + \mathbf{j}x_{ij}$ denote the complex impedance. Let $V_i$ be the complex voltage magnitude at bus $i$, and let $S_{ij} = P_{ij} + \mathbf{j}Q_{ij}$ be the complex power flowing from bus $i$ to bus $j$.

The original non-linear DistFlow equations describing the voltage drop and power balance are:
\begin{align}
    V_j^2 &= V_i^2 - 2(r_{ij}P_{ij} + x_{ij}Q_{ij}) + (r_{ij}^2 + x_{ij}^2)\frac{P_{ij}^2 + Q_{ij}^2}{V_i^2} \label{eq:distflow_v} \\
    P_{ij} &= P_j^{L} + \sum_{k: (j, k) \in \mathcal{E}} \left( P_{jk} + r_{jk}\frac{P_{jk}^2 + Q_{jk}^2}{V_j^2} \right) \label{eq:distflow_p} \\
    Q_{ij} &= Q_j^{L} + \sum_{k: (j, k) \in \mathcal{E}} \left( Q_{jk} + x_{jk}\frac{P_{jk}^2 + Q_{jk}^2}{V_j^2} \right) \label{eq:distflow_q}
\end{align}
where $P_j^L$ and $Q_j^L$ represent the active and reactive load consumption at node $j$.

The LinDistFlow model introduces the assumption of negligible line losses. Particularly, since branch flows in p.u. are typically small, their squares are even smaller compared to the linear voltage drop components. Multiplying this by impedance squared ($r^2 + x^2$), which is also small, makes the value negligible. As a result, the quadratic terms containing $(P^2 + Q^2)$ in Equations \ref{eq:distflow_v}- \ref{eq:distflow_q} are ignored. Applying these simplifications yields the linear update rule for the squared voltage magnitudes:

\begin{equation}
    V_j^2 \approx V_i^2 - 2(r_{ij}P_{ij} + x_{ij}Q_{ij})
    \label{eq:lindistflow_mag}
\end{equation}

\textbf{Implementation:}
We calculate the baseline voltages using a two-step iterative traversal of the radial tree, equivalent to a non-iterative Forward-Backward Pass:
\begin{enumerate}
    \item \textbf{Backward Pass (Power Accumulation):} Traversing from the leaf nodes up to the root (slack bus), we aggregate the loads to determine the branch flows. For a node $j$ fed by node $i$:
    \begin{equation}
        P_{ij} = P_j^{L} + \sum_{k: (j, k) \in \mathcal{E}} P_{jk}, \quad Q_{ij} = Q_j^{L} + \sum_{k: (j, k) \in \mathcal{E}} Q_{jk}
    \end{equation}
    \item \textbf{Forward Pass (Voltage Calculation):} Starting from the slack bus (where $V_{slack}$ is fixed), we traverse downstream to the leaves. We update the voltage of each child node $j$ using the voltage of its parent $i$ and the branch flows calculated in the previous step, via Equation \ref{eq:lindistflow_mag}.
\end{enumerate}

\subsubsection{Voltage Angle Recovery}
While the original Baran and Wu formulation focused on voltage magnitudes and losses, the voltage angles can be recovered using a complementary linearization often utilized in convex relaxations of Optimal Power Flow, as detailed by \citet{farivar2013branch}, and a similar set of common assumptions.

Starting from the voltage drop equation $V_j = V_i - (r_{ij} + \mathbf{j}x_{ij})I_{ij}$ and applying the small-angle approximation ($\sin(\theta_{ij}) \approx \theta_i - \theta_j$ and $\cos(\theta_{ij}) \approx 1$), the voltage angle difference is decoupled from the voltage magnitude. The resulting linear relationship is:
\begin{equation}
    \theta_j \approx \theta_i - \frac{x_{ij}P_{ij} - r_{ij}Q_{ij}}{V_i^2}
    \label{eq:lindistflow_angle}
\end{equation}
where $\theta$ is the voltage angle in radians and $V_i$ is the voltage magnitude of the sending bus $i$.

\textbf{Implementation:}
Using the same branch flows $P_{ij}$ and $Q_{ij}$ computed during the Backward Pass, we perform a Forward Pass to update the angles. Note that unlike the magnitude update which depends on $rP + xQ$, the angle update is driven by the cross-coupling term $xP - rQ$.

\end{document}